\documentclass{article} 
\usepackage{iclr2027_conference,times}
\usepackage{array}
\usepackage{colortbl}
\newcolumntype{G}{!{\color{black!30}\vrule width 0.6pt}}

\usepackage[utf8]{inputenc}
\usepackage[T1]{fontenc}
\usepackage{graphicx}
\usepackage{float}
\usepackage{placeins}

\usepackage{amsmath,amssymb,amsfonts}
\usepackage{booktabs}
\usepackage{microtype}
\usepackage{hyperref}
\usepackage{url}

\title{Beyond Pairwise Attention: Higher-Order Modular Attention for Efficient Sequence Learning
\thanks{Preprint. Under review.}}

\iclrfinalcopy
\author{
Shirin Amiraslani \\
University of Waterloo \\
\texttt{s2amiras@uwaterloo.ca}
\And
Xin Gao \\
York University \\
\texttt{xingao@yorku.ca}
}

\begin{document}
\maketitle
\lhead{}
\begin{abstract}
Sequence modeling tasks can involve intrinsic higher-order dependencies, while
standard self-attention assigns scores to token pairs and does not explicitly
parameterize such interactions. We introduce \textbf{H}igher-\textbf{O}rder
\textbf{M}odular \textbf{A}ttention (\textsc{HOMA}), which fuses pairwise
attention with an explicit triadic attention pathway made tractable through
overlapping blocks, local windows, and a low-rank projection. We compare
\textsc{HOMA} with matched pairwise and purely triadic baselines on controlled
\textsc{parity} and \textsc{match3} tasks, as well as TAPE benchmarks.
\textsc{HOMA} is competitive with or outperforms the baselines, with its
clearest advantages when the underlying dependencies extend beyond the
explicitly modeled triadic order. These advantages are accompanied in several
settings by faster convergence and improved parameter efficiency, with the
learned nonlinear fusion providing an effective mechanism for combining the
pairwise and triadic representations. Overall, our results provide empirical
evidence that \textsc{HOMA} is an effective attention design when task
structure extends beyond pairwise interactions.
\end{abstract}
\section{Introduction}
Transformer architectures have become a standard backbone for sequence
learning and form the foundation of modern large language models
\citep{vaswani2017attention}. In standard self-attention, the score between tokens $i$ and $j$ is given by the dot product of their query and key projections, making the scoring mechanism inherently pairwise. While this formulation has been remarkably
effective, recent theory has identified settings that expose representational limitations of pairwise
attention when modeling higher-order dependencies.

One such limitation concerns how efficiently pairwise attention can communicate joint information across
multiple tokens. \citet{sanford2024representational} formalize a sharp distinction
through two sequence-to-sequence tasks that differ only in arity. In
\textsc{match2}, a token must determine whether it participates in a pair whose
values sum to zero modulo $M$; \textsc{match3} asks the analogous question for
a triple. They show that, in the one-layer setting, solving the inherently
three-way \textsc{match3} task with standard pairwise attention requires model
resources to grow with sequence length, whereas a single third-order attention unit can
solve it with constant embedding dimension. \citet{kozachinskiy2025strassen}
further show that this limitation persists even with arbitrarily high numerical
precision, and establish analogous one-layer lower bounds for function
composition \citep{peng2024composition} and binary relation composition. 

A complementary limitation concerns sensitivity to individual inputs.
\citet{hahn2020} shows that, under fixed-size self-attention, changing a single
input symbol changes the encoder output by only $O(1/N)$ as the sequence grows.
This creates a difficulty for globally sensitive functions such as \textsc{parity},
whose output changes whenever any
single bit is flipped.
\citet{chiang2022overcoming} later showed that \textsc{parity} is representable
at arbitrary length, but their construction is highly parameter-sensitive and
standard Transformers failed to learn it across the settings they tested.
Consistent with this, \citet{bhattamishra2020} report substantial difficulties
in learning and length-generalizing \textsc{parity} with Transformers. These concerns motivate an empirical question: when a target depends on
genuinely higher-order structure, how readily can pairwise attention learn that
dependency under finite model and optimization budgets?

Mechanisms that move beyond pairwise attention have appeared in several
forms, yet increasing interaction order introduces its own computational
trade-offs. The 2-simplicial Transformer \citep{clift2020logic} combines pairwise and
triadic attention, but controls the cubic cost of the latter by restricting its
partner pair to a small set of learned virtual entities, yielding a global,
content-based approximation. Higher-order tensor attention
\citep{sanford2024representational} generalizes attention to $s$-way
multilinear interactions, but is introduced primarily as a theoretical
construction because dense higher-order attention scales rapidly in cost.
Strassen attention \citep{kozachinskiy2025strassen} instead scores triples
through a structured sum of pairwise inner products that permits sub-cubic
evaluation. Hypergraph attention \citep{bai2021hypergraph} captures
higher-order relations over an explicitly constructed hypergraph, while poly-attention
\citep{chakrabarti2026poly} introduces a general framework for higher-order
attention that encompasses standard, tensor, and Strassen attention as special
cases. Collectively, these works
demonstrate that explicit higher-order interactions are feasible, but typically
trade increased interaction order against computation, restricted interaction
structure, or both.

Nevertheless, empirical evidence for higher-order attention on natural
sequence data remains comparatively limited, with much of the literature centered on
theoretical analyses or controlled and synthetic settings
\citep{sanford2024representational,clift2020logic,kozachinskiy2025strassen}.
One natural domain in which higher-order attention may offer an advantage is
protein sequence modeling, as both protein structure and function exhibit
dependencies beyond additive or pairwise effects. For Secondary Structure,
$\beta$-turns impose cooperative backbone constraints across three or more
residues \citep{venkatachalam1968,smith1980}, while helix-terminal
stabilization can involve coordinated interactions among several residues
\citep{chakrabartty1993}. The relevance of multi-residue structure is also reflected in
AlphaFold2, whose Evoformer repeatedly applies triangle-based operations to
residue-pair representations \citep{jumper2021highly}. Contact Prediction provides a complementary motivation, as residue covariation
may be mediated by other positions rather than reflect direct physical contact
\citep{morcos2011direct}, making third-residue context potentially useful for
distinguishing direct from indirect associations.
Finally, at the functional level, epistasis makes the effect of a substitution
depend on its genetic background
\citep{weinreich2013should,poelwijk2016context}; additive-plus-pairwise models
can therefore be insufficient \citep{horovitz1990strategy,wells1990additivity},
with substantial three-way and higher-order components observed across
genotype--phenotype landscapes
\citep{poelwijk2019learning,sailer2017high,wu2016adaptation}.

We introduce \textsc{HOMA}, a Higher-Order Modular Attention mechanism designed as a
drop-in replacement for the self-attention sublayer of standard Transformer
backbones. Rather than replacing pairwise attention, \textsc{HOMA} augments it with an
explicit triadic pathway and combines the two representations through learned fusion. Drawing on efficient pairwise-attention implementations
\citep{beltagy2020longformer,zaheer2020bigbird,shen2018biblostan,qiu2020blockbert}, \textsc{HOMA}'s computation is localized through overlapping
blocks and local triadic windows, with a low-rank projection limiting additional
parameter growth. Our central question is whether making three-way interactions explicit and
combining them with pairwise attention yields a more learnable and
parameter-efficient attention design under controlled computational budgets,
rather than whether pairwise Transformers are fundamentally incapable of
representing such dependencies. We evaluate this question against pairwise
baselines and a purely triadic variant within a modular pipeline with matched
architectures, training protocols, and efficiency profiling. We first use
controlled \textsc{parity} and \textsc{match3} experiments to probe interaction
order, capacity, and learnability, and then evaluate whether these controlled advantages extend to TAPE Secondary Structure, Contact Prediction, and
Fluorescence. Across controlled tasks, \textsc{HOMA} shows its clearest advantages when the
underlying dependencies extend beyond pairwise interactions, with gains in
accuracy, convergence, and parameter efficiency across several settings.
On protein sequence benchmarks, these advantages are task-dependent: \textsc{HOMA}
shows its clearest gains on Secondary Structure and Contact Prediction, while
remaining competitive on Fluorescence.

\section{Method}
\begin{figure}[!h]
    \centering
    \includegraphics[width=0.85\linewidth]{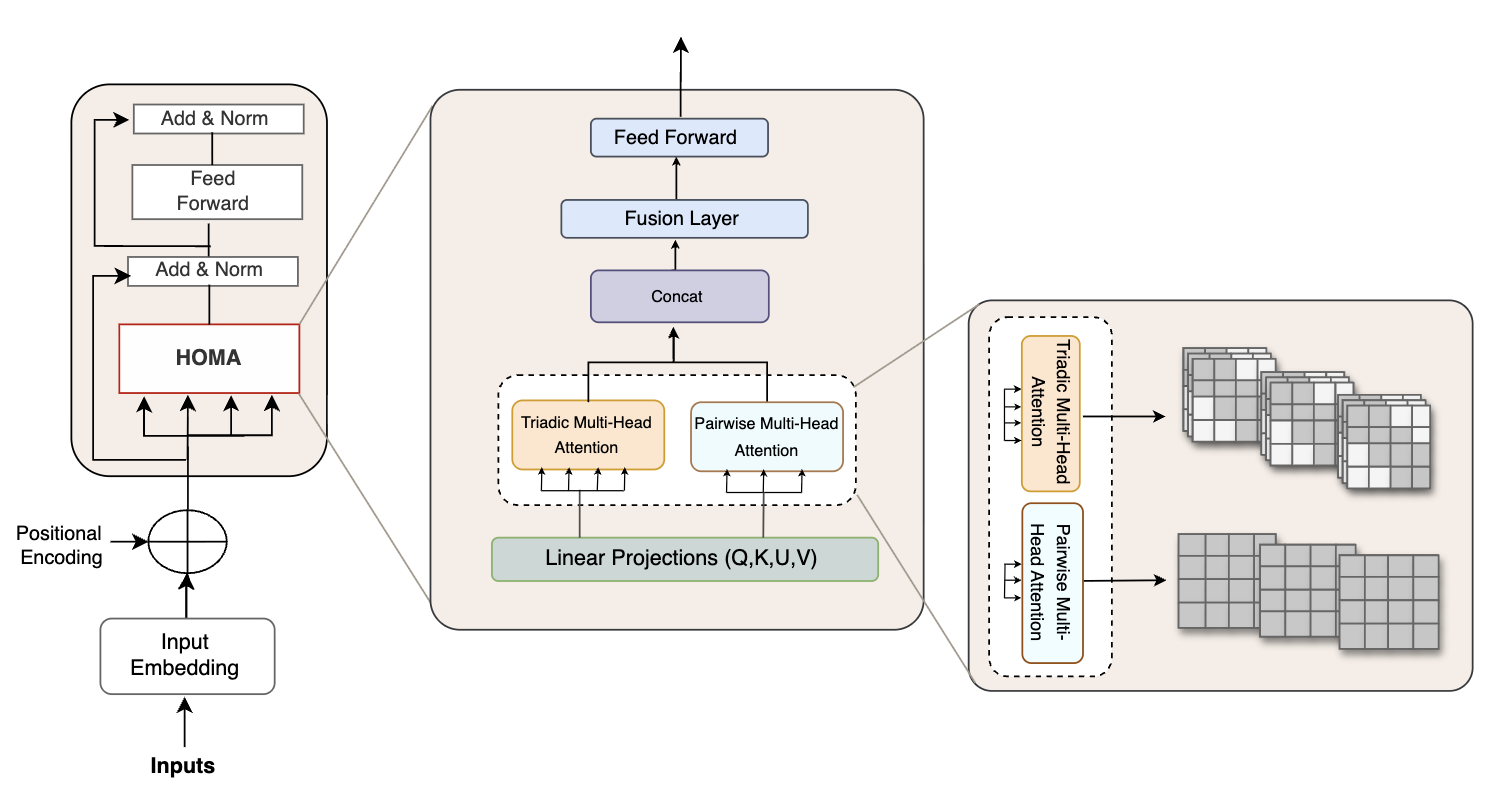}
    \caption{Transformer layer with higher-order modular attention. \textsc{HOMA} replaces the self-attention sublayer and computes pairwise and triadic multi-head attention in parallel from shared projections $Q$, $K$, $V$, and $U$. Their outputs are concatenated, fused, and mapped through the standard output projection.}
    \label{fig:HOMA_architecture}
\end{figure}
As shown in Figure~\ref{fig:HOMA_architecture}, \textsc{HOMA} augments a standard pairwise attention backbone with a triadic interaction pathway. Let $X \in \mathbb{R}^{L \times d_{\text{model}}}$ denote a length-$L$
sequence of embeddings. We form the standard projections
\[
Q = X W^{(Q)}, \qquad
K = X W^{(K)}, \qquad
V = X W^{(V)},
\]
and an additional projection for the triadic pathway,
\[
U = X W^{(U)},
\]
where $W^{(Q)}, W^{(K)}, W^{(V)}, W^{(U)} \in \mathbb{R}^{d_{\text{model}}
\times d_{\text{model}}}$ are learned parameters. We use $H$ attention heads
with $d_{\text{head}} = d_{\text{model}}/H$ and present the following
computations for a single head. Superscripts $(2)$ and $(3)$ denote pairwise
and triadic quantities, respectively.

\paragraph{Pairwise attention.}
Let $Q,K,V \in \mathbb{R}^{L \times d_{\text{head}}}$, where $Q_i$ denotes
the $i$-th row and $Q_{ic}$ its $c$-th component. The scaled dot-product
scores are
\[
S^{(2)}_{ij}=\frac{Q_i K_j^\top}{\sqrt{d_{\text{head}}}},
\qquad
S^{(2)}\in\mathbb{R}^{L\times L}.
\]
Normalizing across $j$ yields attention weights
\[
A^{(2)}_{ij}=\frac{\exp(S^{(2)}_{ij})}{\sum_{j'=1}^L \exp(S^{(2)}_{ij'})},
\]
and the pairwise output is
\[
O^{(2)}_i=\sum_{j=1}^{L}A^{(2)}_{ij}V_j,
\qquad
O^{(2)}\in\mathbb{R}^{L\times d_{\text{head}}}.
\]

\paragraph{Triadic attention.}
Let $U\in\mathbb{R}^{L\times d_{\text{head}}}$. The triadic pathway assigns
weights to ordered pairs $(j,z)$ based on query position $i$ via
\[
S^{(3)}_{ijz}
=
\frac{1}{\sqrt{d_{\text{head}}}}
\sum_{c=1}^{d_{\text{head}}} Q_{ic}K_{jc}U_{zc},
\qquad
S^{(3)}\in\mathbb{R}^{L\times L\times L}.
\]
Normalizing across $(j,z)$ yields
\[
A^{(3)}_{ijz}
=
\frac{\exp(S^{(3)}_{ijz})}
     {\sum_{j'=1}^L\sum_{z'=1}^L \exp(S^{(3)}_{ij'z'})}.
\]
For each pair $(j,z)$ we form a quadratic value interaction
\[
\tilde V_{jz}=V_j\odot V_z\in\mathbb{R}^{d_{\text{head}}},
\]
where $\odot$ denotes elementwise multiplication. The triadic output is
\[
O^{(3)}_i=\sum_{j=1}^{L}\sum_{z=1}^{L}A^{(3)}_{ijz}\,\tilde V_{jz},
\qquad
O^{(3)}\in\mathbb{R}^{L\times d_{\text{head}}}.
\]

\paragraph{Fusion of pairwise and triadic pathways.}
For each position $i$, the pairwise and triadic outputs are concatenated and passed through a fusion network
\[
O^{\text{cat}}_{i} = \big[ O^{(2)}_{i} \,\|\, O^{(3)}_{i} \big]
\in \mathbb{R}^{2d_{\text{head}}},
\qquad
O^{\text{fuse}}_{i} = g\!\left(O^{\text{cat}}_{i}\right)
\in \mathbb{R}^{d_{\text{head}}},
\]
where $g(\cdot)$ is a two-layer MLP with ReLU activation that acts as an
adaptive gate, allocating representational capacity between the pairwise and
triadic channels as dictated by the data. Before softmax normalization, masks
are applied to both the pairwise and triadic attention weights. After
concatenating the $H$ head outputs, the standard output projection
$W^{(O)}\in \mathbb{R}^{d_{\text{model}} \times d_{\text{model}}}$ is
applied. 

For comparison, we also consider \textsc{Blockwise-3D}, a purely triadic
mechanism that computes $S^{(3)}$, $A^{(3)}$, and $O^{(3)}$ identically to
\textsc{HOMA}'s triadic pathway; this attention variant omits the pairwise pathway and fusion network and applies $W^{(O)}$ directly to the concatenated triadic head outputs.

\subsection{Efficient implementation}
The triadic pathway is the primary computational challenge; naive
evaluation of $S^{(3)}_{ijz}$ over all triplets scales as $\mathcal{O}(L^3
d_{\text{head}})$ in compute and $\mathcal{O}(L^3)$ in storage, which is
infeasible for typical protein lengths. \textsc{HOMA} and \textsc{Blockwise-3D} address this through
three complementary optimizations: overlapping block decomposition to
localize computation, windowed triadic attention to reduce cubic scaling
within each block, and a low-rank $U$ projection to control parameter
growth.
\paragraph{Overlapping block decomposition.}
Block-granular computation aligns with two
complementary properties: contiguous tiling maximizes GPU memory throughput
and Tensor Core utilization \citep{dao2022flashattention}, and neighboring
keys empirically carry similar attention weights \citep{jiang2024}, so
contiguous blocks capture coherent attention mass without scattering it
across distant positions. We partition the sequence into
overlapping blocks of length $\ell$ with stride $s$, giving
\[
T = \left\lfloor \frac{L-\ell}{s} \right\rfloor + 1
\]
blocks and block outputs are merged back to the full sequence by
overlap-averaging. In \textsc{HOMA}, both pairwise and triadic attentions are computed independently within
each block.

\paragraph{Windowed triadic attention within blocks.}
Within each block, we restrict $(j,z)$ to a local neighborhood of size $w
\ll \ell$ around each query position. This reduces per-block triadic compute
from $\mathcal{O}(\ell^3 d_{\text{head}})$ to
$\mathcal{O}(\ell w^2 d_{\text{head}})$ while preserving local higher-order
interactions.

\paragraph{Low-rank $U$ projection.}
To limit parameter growth in the triadic pathway, we factorize $W^{(U)}$ as
\[
W^{(U)} = W^{(U_u)} W^{(U_v)},
\qquad
W^{(U_u)} \in \mathbb{R}^{d_{\text{model}} \times r}, \quad
W^{(U_v)} \in \mathbb{R}^{r \times d_{\text{model}}}, \quad
r \ll d_{\text{model}}.
\]
This retains the expressivity of an additional projection while substantially
reducing parameters relative to a full $d_{\text{model}} \times d_{\text{model}}$
map.

With $T$ overlapping blocks of length $\ell$, pairwise attention costs
$\mathcal{O}(T\,\ell^2 d_{\text{head}})$ and the windowed triadic pathway
costs $\mathcal{O}(T\,\ell\,w^2 d_{\text{head}})$.

\textsc{HOMA}'s block-and-window design induces a deliberate two-scale structure. Within
each overlapping block, pairwise attention operates over the full $\ell \times
\ell$ token-pair space; triadic attention is further restricted to a local
window of size $w \leq \ell$. The
pairwise pathway therefore operates at a more global scale capturing long-range
dependencies, while the triadic pathway operates at a local scale capturing
short-range higher-order cooperativities, and their combination aims to yield
representations richer than either scale alone.

\section{Experimental Setup}
\label{sec:setup}
All experiments use a shared pipeline in which only the attention mechanism
varies. We compare \textsc{Pairwise-2D}, standard softmax self-attention;
\textsc{Blockwise-2D}, its overlapping-block variant;
\textsc{Blockwise-3D}, a purely triadic mechanism; and \textsc{HOMA}, our proposed pairwise--triadic attention mechanism. All experimental results are averaged across three seeds. Full
hyperparameters and training setup are provided in Appendix~\ref{app:hparams},
Tables~\ref{tab:app_hp_data}--\ref{tab:app_hp_optim}. All experiments run on
a single NVIDIA A100-SXM4-80GB GPU.

\subsection{Synthetic Diagnostic Tasks}

Diagnostic models consist of an embedding layer, the attention mechanism under
study, and a linear output classifier, with the standard post-attention
feed-forward MLP omitted to limit additional nonlinear feature composition. Because
\textsc{HOMA} includes an internal fusion MLP, we also report \textsc{HOMA-add}, which replaces
learned fusion with a parameter-free sum of the pairwise and triadic outputs,
isolating the contribution of learned fusion from that of combining the two
interaction pathways. We consider two synthetic task families, \textsc{parity}--\textsc{majority}
and \textsc{match2}--\textsc{match3}, to probe how interaction order and model
capacity affect learnability under controlled conditions.
\paragraph{\textsc{parity}-$k$ and \textsc{majority}-$k$.}
Given a random binary sequence $b \in \{0,1\}^L$, a set of $k$ offsets
$\mathcal{O}=\{o_1,\dots,o_k\}$, and an interior position $i$, we define
\[
  y^{\textsc{par}}_i
  = \bigoplus_{j=1}^{k} b_{i+o_j},
  \qquad
  y^{\textsc{maj}}_i
  = \mathbf{1}\!\left[
      \textstyle\sum_{j=1}^{k} b_{i+o_j} > k/2
    \right].
\]
Here $\oplus$ denotes XOR.
\textsc{parity}-$k$ is an irreducible $k$-way interaction since every proper
subset of its relevant inputs is statistically independent of the label.
\textsc{majority}-$k$ is a threshold on their additive sum and therefore
serves as a lower-order control. The $k$ offsets are spread evenly over the
fixed interval $[-R,R]$, with $R=3$, holding interaction reach fixed as $k$
varies. Only positions whose full offset pattern lies within the sequence are
labelled.

\paragraph{\textsc{match2} and \textsc{match3}.}
For integer sequences $x \in \mathbb{Z}_M^N$
\citep{sanford2024representational},
\[
  y^{(2)}_i
  = \mathbf{1}\!\left[
      \exists j:\; x_i+x_j\equiv0 \pmod M
    \right],
  \qquad
  y^{(3)}_i
  = \mathbf{1}\!\left[
      \exists j,z:\; x_i+x_j+x_z\equiv0 \pmod M
    \right].
\]
The modulus \(M\) is calibrated separately for each sequence length to maintain approximately balanced labels (Appendix~\ref{app:match_setup}), and each integer is represented using a fixed Fourier basis. The modular-sum condition in \textsc{match3} admits a Fourier representation
whose expansion contains products of three token-dependent coordinates,
aligning naturally with the trilinear structure of triadic attention. We use \textsc{match3} as a controlled probe of explicit
three-way interaction, with \textsc{match2} as the corresponding order-2
control.

For \textsc{parity}, we vary interaction order, embedding size, and depth;
for \textsc{match3}, we vary embedding size using a
single attention layer. The single-layer setting is particularly informative because it removes cross-layer composition and more directly isolates differences between attention mechanisms. On both task families, block size equals sequence
length, so \textsc{Pairwise-2D} and \textsc{Blockwise-2D} are identical. The triadic window covers all task-relevant positions:
$w=2N-1$ for \textsc{match2}/\textsc{match3} and $w=2R+1=7$ for
\textsc{parity}. Thus, receptive-field differences do not confound these
comparisons; locality is examined separately in
Section~\ref{sec:ablations}.

\subsection{TAPE Protein Benchmarks}

We evaluate on three supervised tasks from the TAPE benchmark \citep{rao2019tape}, spanning residue-level and pairwise prediction, and sequence-level regression. Models are trained without pretrained protein representations and use no multiple sequence alignments or coevolutionary features; the TAPE
experiments are therefore designed as controlled comparisons of attention
mechanisms rather than direct state-of-the-art comparisons. 

\paragraph{Secondary Structure.}
Secondary Structure prediction is a residue-level classification task in which
the model assigns each residue one of three structural labels:
$\{\mathrm{helix},\,\mathrm{strand},\,\mathrm{coil}\}$. Training and validation
data are sourced from \citet{netsurfp}, and we evaluate on the held-out CB513
\citep{cuff1999}, CASP12 \citep{casp}, and TS115 \citep{yang2018} test sets.
We report Q3 accuracy over non-padded residues.

\paragraph{Contact Prediction.}
Contact Prediction is a pairwise residue-classification task in which the model
predicts a binary contact matrix $y\in\{0,1\}^{L\times L}$. Following TAPE
\citep{rao2019tape}, two residues are considered in contact if their
C$\alpha$ atoms lie within $8$~\AA{} in the experimentally resolved structure.
Data are drawn from ProteinNet \citep{alquraishi2019}, with evaluation on the
held-out CASP12 split \citep{casp}. We report precision at $L/5$ over
long-range residue pairs ($|i-j|\geq24$).

\paragraph{Fluorescence.}
Fluorescence is a sequence-level regression task over mutated variants of green
fluorescent protein \citep{sarkisyan2016}, with log fluorescence intensity as
the prediction target. Following TAPE \citep{rao2019tape}, training variants
lie within Hamming distance three of the parent sequence, whereas test variants
contain at least four mutations. We report Spearman's rank correlation $\rho$.

\section{Results}
\label{sec:results}

\subsection{Synthetic Diagnostic Tasks}
Table~\ref{tab:parity_order_d64} reports results on \textsc{parity}. All models solve the \textsc{majority} control and achieve perfect or near-perfect accuracy on \textsc{parity} tasks of orders 1--2 (Appendix~\ref{app:parity}, Table~\ref{tab:app_parity_extra}). \textsc{Pairwise-2D} can recover \textsc{parity}-3, but its accuracy degrades sharply on \textsc{parity}-4 and \textsc{parity}-5, and increasing width or depth only partially closes this gap within the training budget. \textsc{Blockwise-3D} performs substantially better on the higher-order tasks and benefits from additional depth, while \textsc{HOMA-add} further improves performance in shallow and lower-dimensional regimes. \textsc{HOMA} with nonlinear fusion achieves perfect accuracy across all tested interaction orders, embedding dimensions, and depths.

\begin{figure}[!h]
    \centering
    \includegraphics[width=1\linewidth]{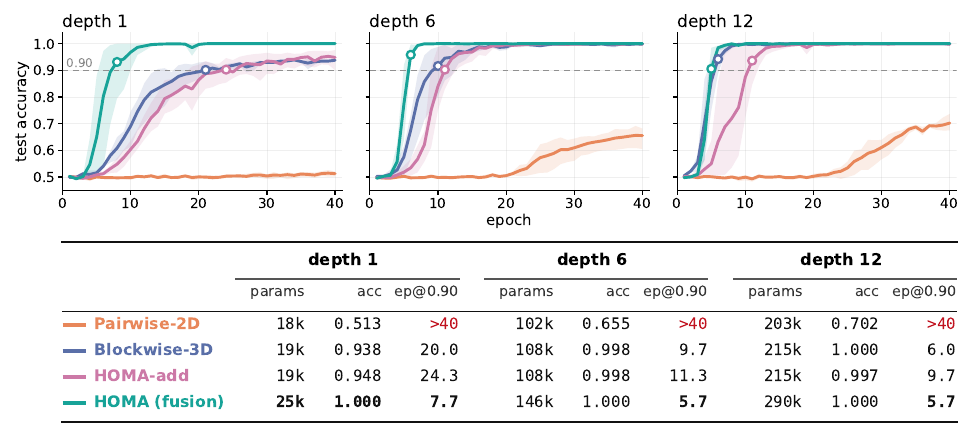}
    \caption{\textbf{Convergence on \textsc{parity}-5 at embedding
size $d=64$ across network depths $D\in\{1,6,12\}$.}
Curves show mean test accuracy over three seeds, with final values
reported as \texttt{acc}; shaded bands span the minimum--maximum
range across seeds.
The dashed line marks $0.90$ accuracy.
Open circles and \texttt{ep@0.90} indicate the mean first-crossing
epoch across seeds; $>40$ indicates that no seed reached the
threshold within 40 epochs.}
\label{fig:parity_convergence}
\end{figure}

\textsc{HOMA}’s benefits on higher-order tasks extend beyond final accuracy to improved
parameter efficiency and faster convergence. On \textsc{parity}-3 with a
single attention layer (Appendix~\ref{app:parity},
Table~\ref{tab:app_parity3_capacity}), both \textsc{Pairwise-2D} and \textsc{HOMA}
achieve perfect accuracy, but \textsc{HOMA} requires approximately $5.4\times$ fewer
parameters and converges in roughly half as many epochs. The separation becomes
more pronounced on \textsc{parity}-5
(Figure~\ref{fig:parity_convergence}): \textsc{HOMA} solves the task already at depth
$1$, whereas \textsc{Pairwise-2D} fails within the training budget at every tested depth,
and \textsc{Blockwise-3D} reaches perfect accuracy at depth $12$. Notably, with a single
layer, \textsc{HOMA-add} still outperforms both baselines in final accuracy but requires
about three times as many epochs as \textsc{HOMA} to reach $0.90$ accuracy. These results suggest that combining pairwise and triadic representations is advantageous for higher-order interaction learning, while their nonlinear fusion provides a particularly efficient mechanism for faster convergence when estimating dependencies beyond the primitive interaction order.

%
%

\definecolor{muted}{HTML}{A6ACB2}
\newcommand{\sd}[1]{\,{\tiny\textcolor{black!55}{$\pm #1$}}}
\begin{table}[!h]
\centering
\scriptsize
\setlength{\tabcolsep}{0.7pt}
\caption{\textbf{\textsc{parity}-$k$ across interaction order $k$, network
depth $D\in\{1,6,12\}$, and embedding size $d$.}
Mean final test accuracy $\pm$ standard deviation over three seeds. Results for $k=3$ are shown in gray since performance is near-saturated across attention mechanisms. \textsc{HOMA} results for $k=4,5$ are bold.}
\label{tab:parity_order_d64}

\begin{tabular}{@{}ll ccc G ccc G ccc@{}}
\toprule
& & \multicolumn{3}{c}{\textbf{$k{=}3$}}
  & \multicolumn{3}{c}{\textbf{$k{=}4$}}
  & \multicolumn{3}{c}{\textbf{$k{=}5$}} \\
\cmidrule(lr){3-5}
\cmidrule(lr){6-8}
\cmidrule(lr){9-11}

$d$ & \textbf{mechanism}
& $D{=}1$ & $6$ & $12$
& $D{=}1$ & $6$ & $12$
& $D{=}1$ & $6$ & $12$ \\
\midrule

$\mathbf{32}$
& \textsc{Pairwise-2D}
& \textcolor{muted}{.978}\sd{.037}
& \textcolor{muted}{1.000}\sd{.000}
& \textcolor{muted}{1.000}\sd{.000}
& .499\sd{.014}
& .729\sd{.021}
& .694\sd{.011}
& .503\sd{.002}
& .611\sd{.036}
& .601\sd{.028} \\

& \textsc{Blockwise-3D}
& \textcolor{muted}{1.000}\sd{.000}
& \textcolor{muted}{1.000}\sd{.000}
& \textcolor{muted}{1.000}\sd{.000}
& .964\sd{.048}
& .997\sd{.003}
& 1.000\sd{.000}
& .847\sd{.125}
& .991\sd{.011}
& 1.000\sd{.001} \\

& \textsc{HOMA-add}
& \textcolor{muted}{1.000}\sd{.000}
& \textcolor{muted}{1.000}\sd{.000}
& \textcolor{muted}{1.000}\sd{.000}
& .992\sd{.010}
& 1.000\sd{.000}
& 1.000\sd{.000}
& .871\sd{.066}
& .996\sd{.007}
& .999\sd{.002} \\

& \textsc{HOMA}
& \textcolor{muted}{1.000}\sd{.000}
& \textcolor{muted}{1.000}\sd{.000}
& \textcolor{muted}{1.000}\sd{.000}
& \textbf{1.000}\sd{.000}
& \textbf{1.000}\sd{.000}
& \textbf{1.000}\sd{.000}
& \textbf{1.000}\sd{.000}
& \textbf{1.000}\sd{.000}
& \textbf{1.000}\sd{.000} \\

\addlinespace[3pt]
\midrule

$\mathbf{64}$
& \textsc{Pairwise-2D}
& \textcolor{muted}{1.000}\sd{.000}
& \textcolor{muted}{1.000}\sd{.000}
& \textcolor{muted}{1.000}\sd{.000}
& .500\sd{.004}
& .782\sd{.101}
& .728\sd{.039}
& .513\sd{.010}
& .655\sd{.043}
& .702\sd{.032} \\

& \textsc{Blockwise-3D}
& \textcolor{muted}{1.000}\sd{.000}
& \textcolor{muted}{1.000}\sd{.000}
& \textcolor{muted}{1.000}\sd{.000}
& .979\sd{.018}
& 1.000\sd{.000}
& 1.000\sd{.000}
& .938\sd{.001}
& .998\sd{.003}
& 1.000\sd{.000} \\

& \textsc{HOMA-add}
& \textcolor{muted}{1.000}\sd{.000}
& \textcolor{muted}{1.000}\sd{.000}
& \textcolor{muted}{1.000}\sd{.000}
& .992\sd{.011}
& 1.000\sd{.000}
& 1.000\sd{.000}
& .948\sd{.041}
& .998\sd{.003}
& .997\sd{.005} \\

& \textsc{HOMA}
& \textcolor{muted}{1.000}\sd{.000}
& \textcolor{muted}{1.000}\sd{.000}
& \textcolor{muted}{1.000}\sd{.000}
& \textbf{1.000}\sd{.000}
& \textbf{1.000}\sd{.000}
& \textbf{1.000}\sd{.000}
& \textbf{1.000}\sd{.000}
& \textbf{1.000}\sd{.000}
& \textbf{1.000}\sd{.000} \\

\bottomrule
\end{tabular}
\end{table}
\begin{table}[!h]
\centering\footnotesize\setlength{\tabcolsep}{3.4pt}
\caption{\textbf{\textsc{match3} across sequence lengths $N$
and embedding sizes $d_{\mathrm{model}}$.}
Mean final test accuracy $\pm$ standard deviation over three seeds;
$M$ denotes the modulus.
Bold marks the highest mean accuracy at each length and embedding size.
$d_{0.90}$ is the smallest tested embedding size achieving mean
accuracy $\geq 0.90$.
params reports the parameter count at $d_{0.90}$ relative
to \textsc{Pairwise-2D} at the same embedding size.
A dash means the threshold is never reached within the tested range,
also indicated by $d_{0.90}>256$.
  The two additional widths $d_{\mathrm{model}}=8$ and
$256$ are in Appendix~\ref{app:match}, Table~\ref{tab:app_match3}.}
\label{tab:match3}
\begin{tabular*}{\linewidth}{@{\extracolsep{\fill}}ll cccc G cc@{}}
\toprule
& & \multicolumn{4}{c}{test accuracy at $d_{\mathrm{model}}$} & & \\
\cmidrule(lr){3-6}
$N\,(M)$ & mechanism & $16$ & $32$ & $64$ & $128$ & $d_{0.90}$ & params \\
\midrule
$6\,(30)$ & \textsc{Pairwise-2D} & 0.816\,{\scriptsize$\pm$0.009} & 0.835\,{\scriptsize$\pm$0.001} & 0.830\,{\scriptsize$\pm$0.003} & 0.783\,{\scriptsize$\pm$0.009} & $>$256 & -- \\
 & \textsc{Blockwise-3D} & 0.836\,{\scriptsize$\pm$0.007} & 0.921\,{\scriptsize$\pm$0.012} & 0.958\,{\scriptsize$\pm$0.008} & 0.927\,{\scriptsize$\pm$0.028} & $32$ & 1.08$\times$ \\
 & \textsc{HOMA-add} & 0.835\,{\scriptsize$\pm$0.003} & 0.929\,{\scriptsize$\pm$0.002} & 0.990\,{\scriptsize$\pm$0.002} & 0.975\,{\scriptsize$\pm$0.006} & $32$ & 1.08$\times$ \\
 & \textsc{HOMA} & \textbf{0.857}\,{\scriptsize$\pm$0.006} & \textbf{0.967}\,{\scriptsize$\pm$0.007} & \textbf{0.993}\,{\scriptsize$\pm$0.002} & \textbf{0.991}\,{\scriptsize$\pm$0.003} & $32$ & 1.60$\times$ \\
\addlinespace[4pt]
$8\,(56)$ & \textsc{Pairwise-2D} & 0.708\,{\scriptsize$\pm$0.005} & 0.744\,{\scriptsize$\pm$0.004} & 0.761\,{\scriptsize$\pm$0.009} & 0.731\,{\scriptsize$\pm$0.012} & $>$256 & -- \\
 & \textsc{Blockwise-3D} & 0.726\,{\scriptsize$\pm$0.013} & 0.768\,{\scriptsize$\pm$0.005} & \textbf{0.856}\,{\scriptsize$\pm$0.004} & 0.912\,{\scriptsize$\pm$0.004} & $128$ & 1.03$\times$ \\
 & \textsc{HOMA-add} & 0.734\,{\scriptsize$\pm$0.014} & 0.776\,{\scriptsize$\pm$0.004} & 0.848\,{\scriptsize$\pm$0.003} & 0.921\,{\scriptsize$\pm$0.004} & $128$ & 1.03$\times$ \\
 & \textsc{HOMA} & \textbf{0.745}\,{\scriptsize$\pm$0.009} & \textbf{0.799}\,{\scriptsize$\pm$0.011} & 0.853\,{\scriptsize$\pm$0.003} & \textbf{0.925}\,{\scriptsize$\pm$0.008} & $128$ & 1.19$\times$ \\
\bottomrule
\end{tabular*}
\end{table}

Table~\ref{tab:match3} reports results on \textsc{match3}. Under the same
protocol, \textsc{match2} achieves $1.000$ accuracy over three seeds at every
displayed width, except for one \textsc{Pairwise-2D} seed at $N=8$,
$d_{\mathrm{model}}=128$ (Appendix~\ref{app:match},
Table~\ref{tab:app_match2}). On \textsc{match3},
increasing width yields relatively limited and non-monotonic gains for
\textsc{Pairwise-2D}, whose best accuracy remains below $0.90$. In contrast, all
mechanisms with explicit triadic interactions cross $0.90$ at $d=32$ for
$N=6$ and $d=128$ for $N=8$. \textsc{HOMA} attains the highest overall accuracy in
both settings, reaching $0.993$ and $0.925$ for $N=6$ and $N=8$,
respectively.
\subsection{TAPE Protein Benchmarks}
\begin{figure}[!h]
    \centering
    \includegraphics[width=0.9\linewidth]{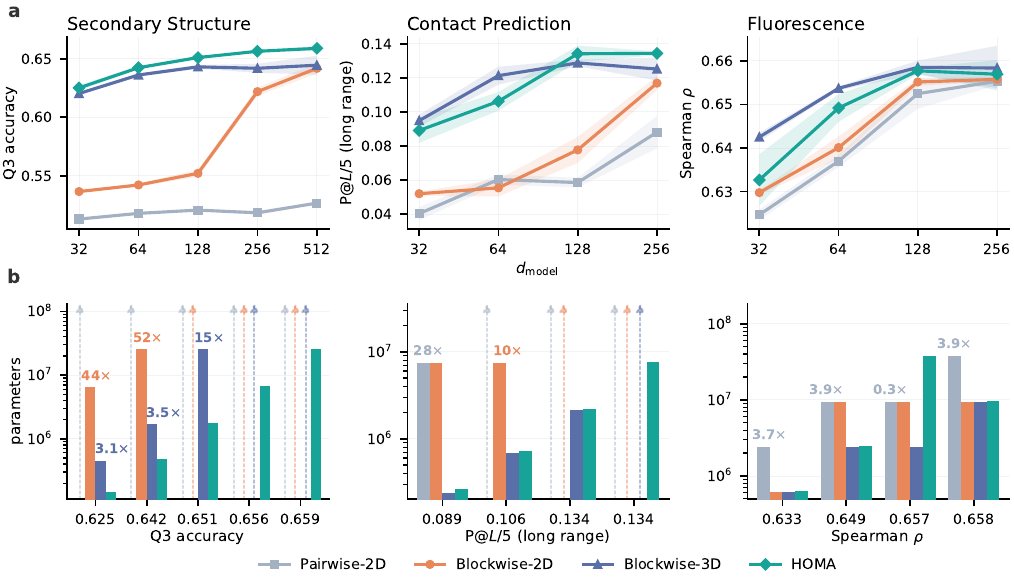}
\caption{\textbf{Test performance and performance-matched parameter efficiency
across TAPE Secondary Structure (CB513), Contact Prediction, and
Fluorescence.}
Results are averaged over three seeds.
\textbf{(a)} Mean performance across embedding sizes, with shaded regions
showing $\pm1$ standard deviation.
\textbf{(b)} For each attention mechanism, we report the minimum parameter
count across tested widths that matches each \textsc{HOMA} performance target.
A run is considered to match a target if its score is within $1.5$ pooled
seed standard deviations of the \textsc{HOMA} score or within $0.5\%$ in relative
terms. Dotted arrows denote targets not reached; annotations show parameter
count relative to \textsc{HOMA}. The parameter axis is logarithmic.}
\label{fig:tape_grid}
\end{figure}

Figure~\ref{fig:tape_grid}a summarizes test performance for Secondary Structure
(CB513), Contact Prediction, and Fluorescence. \textsc{Blockwise-2D} tends to improve over \textsc{Pairwise-2D} across all three TAPE tasks. \textsc{HOMA} shows its clearest gains on Secondary Structure and Contact
Prediction, with the gap over pairwise attention diminishing at the largest
tested widths. On Secondary Structure, \textsc{HOMA} outperforms both pairwise and
purely triadic baselines across all tested embedding sizes. On Contact
Prediction, \textsc{Blockwise-3D} is stronger at smaller widths, while \textsc{HOMA} overtakes it
at $d_{\mathrm{model}}\geq128$. Fluorescence shows substantially weaker
separation, with \textsc{Blockwise-3D} slightly outperforming \textsc{HOMA} overall. Full results across TAPE
evaluations, including all Secondary Structure test sets, are provided in
Appendix~\ref{app:tape}, Table~\ref{tab:app_tape_tests}.

Figure~\ref{fig:tape_grid}b compares the parameter count required to match \textsc{HOMA}'s performance at
each embedding size. On Secondary Structure, the strongest pairwise baseline
requires up to $52\times$ more parameters to match \textsc{HOMA}, while \textsc{Blockwise-3D}
requires up to $15\times$ more parameters. On Contact Prediction, the advantage is
primarily over pairwise attention, which requires up to $28\times$ more
parameters. Full scores by width and the corresponding throughput and peak
memory are reported in Appendices~\ref{app:tape} and~\ref{app:memory}.
\textsc{HOMA} incurs greater memory cost at a fixed embedding size because of its triadic pathway, but this overhead can be offset in some matched-performance settings: on
Secondary Structure, \textsc{HOMA} at $d_{\mathrm{model}}=64$ uses approximately
$2.97$\,GB, whereas matching it requires \textsc{Blockwise-2D} at
$d_{\mathrm{model}}=512$ ($3.55$\,GB) and \textsc{Blockwise-3D} at
$d_{\mathrm{model}}=128$ ($3.11$\,GB). Full peak-memory results at every width
are provided in Appendix~\ref{app:memory}, Table~\ref{tab:app_cost_all}.

\section{Ablations and Analysis}
\label{sec:ablations}
\begin{figure}[t]
\centering
\includegraphics[width=0.95\linewidth]{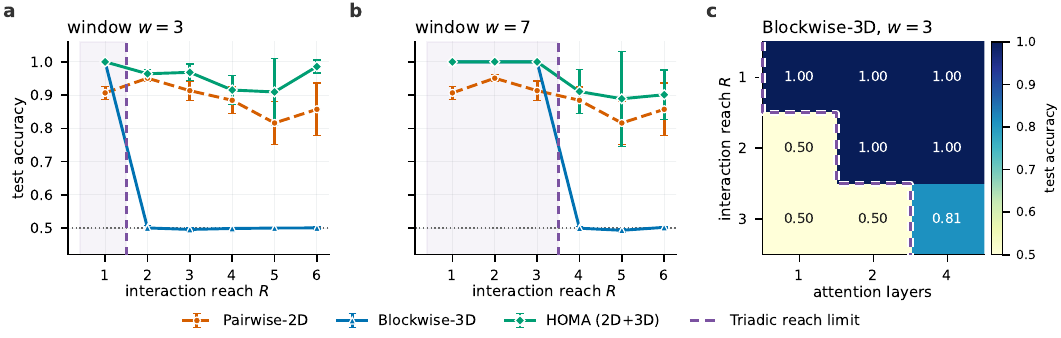}
\caption{\textbf{Triadic window coverage on \textsc{parity}-3.}
All panels use dedicated coverage-sweep runs with $d_{\mathrm{model}}=32$ and
sequence length $L=24$; results are averaged over three seeds.
\textbf{(a,b)} Test accuracy across interaction reach $R$ for window sizes
$w{=}3$ and $w{=}7$; the shaded region and dashed line indicate the triadic
reach limit $R_w$.
\textbf{(c)} \textsc{Blockwise-3D} accuracy across interaction reach and depth for
$w{=}3$.}
\label{fig:coverage_sweep}
\end{figure}

We conduct ablations on \textsc{parity}-$3$ to examine how window size and
depth affect the receptive field of the triadic pathway. We define the
interaction reach $R$ as the maximum absolute offset from the query position
among the positions entering the \textsc{parity}-$3$ interaction, and the
triadic window reach as $R_w=(w{-}1)/2$, so a window of size $w$ covers
offsets up to $\pm R_w$. Accordingly,
Figure~\ref{fig:coverage_sweep}(a,b) shows that \textsc{Blockwise-3D} achieves perfect
accuracy when $R\leq R_w$ but drops to chance beyond this range. \textsc{HOMA} avoids this
collapse because its pairwise branch remains available, maintaining or
exceeding the \textsc{Pairwise-2D} baseline. Depth partially mitigates this limitation,
as shown in Figure~\ref{fig:coverage_sweep}(c): $R=2$ is solved at depth 2,
while $R=3$ only partially recovers at depth 4. Overall, the triadic pathway
degrades when the interaction reach exceeds its local window reach, although
additional depth can recover some longer-range dependencies. The corresponding
numerical results are provided in Appendix~\ref{app:coverage},
Tables~\ref{tab:app_coverage_cliff}--\ref{tab:app_coverage_depth}.

\begin{table}[!h]
\centering\footnotesize\setlength{\tabcolsep}{3.0pt}
\caption{\textbf{Component teardown of \textsc{HOMA} on \textsc{parity}-$k$ and
\textsc{match3}.}
Each row ablates one component with the others fixed.
Results use one attention layer, $d_{\mathrm{model}}{=}32$ and three seeds.
acc is mean final test accuracy; ep@0.90 is the mean epoch to
reach $0.90$ accuracy among successful seeds within 40 epochs.
$^\dagger$ and $^\ddagger$ denote one and two successful seeds, respectively;
$>40$ denotes no successful seed.}
\label{tab:teardown}
{\fontsize{7.5}{9}\selectfont
\begin{tabular}{@{}l ccc |rr |rr |rr |rr@{}}
\toprule
 & \multicolumn{3}{c}{Component} & \multicolumn{2}{c}{\textsc{parity}-3} & \multicolumn{2}{c}{\textsc{parity}-4} & \multicolumn{2}{c}{\textsc{parity}-5} & \multicolumn{2}{c}{\textsc{match3}} \\
\cmidrule(lr){2-4}\cmidrule(lr){5-6}\cmidrule(lr){7-8}\cmidrule(lr){9-10}\cmidrule(lr){11-12}
Design & pooling & third factor & fusion & acc & ep@0.90 & acc & ep@0.90 & acc & ep@0.90 & acc & ep@0.90 \\
\midrule
\textbf{\textsc{HOMA}} & \textbf{softmax} & \textbf{$W^{(U)}$} & \textbf{MLP} & \textbf{1.000} & \textbf{5.7} & \textbf{1.000} & \textbf{7.7} & \textbf{1.000} & \textbf{9.3} & \textbf{0.967} & \textbf{10.7} \\
\midrule
Replace & uniform & $W^{(U)}$ & MLP & 1.000 & 9.7 & 0.994 & 16.3 & 0.955 & 21.7 & 0.861 & $>$40 \\
 & softmax & $W^{(U)}{=}W^{(K)}$ & MLP & 1.000 & 4.3 & 1.000 & 7.3 & 0.983 & 9.7 & 0.933 & 19.3 \\
\midrule
Replace & softmax & $W^{(U)}$ & sum & 1.000 & 7.3 & 0.992 & 17.0 & 0.871 & 32.0$^{\dagger}$ & 0.929 & 19.3 \\
 & uniform & $W^{(U)}$ & sum & 0.988 & 18.7 & 0.495 & $>$40 & 0.500 & $>$40 & 0.838 & $>$40 \\
 & softmax & $W^{(U)}{=}W^{(K)}$ & sum & 1.000 & 6.3 & 0.978 & 23.3 & 0.909 & 25.0$^{\ddagger}$ & 0.904 & 32.0 \\
\bottomrule
\end{tabular}}
\end{table}

Table~\ref{tab:teardown} isolates three design choices in \textsc{HOMA}. We remove
selective triadic attention by replacing it with uniform pooling, remove the
independent parameterization of the third projection by equating $W^{(U)}$
with $W^{(K)}$, and remove the learned fusion MLP by replacing it with
parameter-free addition. Removing the fusion layer substantially slows
convergence, especially on \textsc{parity}-$4$ and \textsc{parity}-$5$, with
accuracy also dropping. With the fusion MLP retained, uniform pooling mainly
affects convergence speed, with accuracy also decreasing on \textsc{match3}.
The distinction becomes clearer without the fusion layer, where uniform pooling
reduces \textsc{parity}-$4$ and \textsc{parity}-$5$ to chance, matching the
pairwise baseline, while also substantially slowing convergence. Tying $U$ to
$K$ has a smaller and less consistent effect across settings. Overall, the
fusion layer is beneficial for faster convergence, while selective triadic
attention provides the clearest contribution to \textsc{HOMA} on higher-order tasks.

\section{Conclusion}
\label{sec:conclusion}

We present \textsc{HOMA}, a Higher-Order Modular Attention mechanism that combines
pairwise and triadic pathways with localized higher-order computation to make
triadic attention tractable. Across controlled experiments, \textsc{HOMA} becomes
increasingly effective as interaction order grows, with learned nonlinear
fusion further improving convergence. On TAPE, these advantages extend to
natural sequence data in a task-dependent manner. Overall, the results support
\textsc{HOMA} as a useful inductive bias when task structure extends beyond pairwise
interactions. Future work will address the computational and locality
constraints of the triadic pathway and evaluate \textsc{HOMA} in pretrained models and
broader sequence domains.

\subsection*{AI use statement}
Generative AI tools were used for writing assistance, including copy-editing,
rephrasing, and \LaTeX{} formatting of text drafted by the authors, and to assist
with code implementation and debugging. All AI-assisted text and code were
reviewed, tested, and verified by the authors. The authors take full
responsibility for the final content of this work, including the implementation,
reported results, text, claims, and released artifacts.

\subsection*{Ethics statement}
This work uses only publicly released protein benchmark datasets distributed through TAPE
\citep{rao2019tape} and involves no human subjects, personally identifying information, or
sensitive data. The released models are small-scale attention-mechanism prototypes trained
on public benchmarks and pose no foreseeable risk of misuse. We are aware of no conflicts
of interest associated with this work.

\subsection*{Reproducibility statement}
All architectural, optimizer, and attention-specific hyperparameters are reported in
Section~\ref{sec:setup} and in full in Appendix~\ref{app:hparams},
Tables~\ref{tab:app_hp_data}--\ref{tab:app_hp_optim}.
Section~\ref{sec:setup} specifies the dataset splits,
tokenization, evaluation metrics, and profiling methodology for all three tasks. All
results are reported as mean $\pm$ standard deviation over three fixed random seeds, run on a single NVIDIA A100-SXM4-80GB GPU. An anonymized code package
containing model implementations and training and evaluation scripts, together with
instructions for reproducing the reported experiments,
is described in Appendix~\ref{app:code}.

\bibliography{references}

@inproceedings{rao2019tape,
  author    = {Rao, Roshan and Bhattacharya, Nicholas and Thomas, Neil and Duan, Yan and Chen, Xi and Canny, John and Abbeel, Pieter and Song, Yun S.},
  title     = {Evaluating Protein Transfer Learning with {TAPE}},
  booktitle = {Advances in Neural Information Processing Systems},
  year      = {2019}
}

@article{venkatachalam1968,
  author    = {Venkatachalam, C. M.},
  title     = {Stereochemical criteria for polypeptides and proteins.
               {V}. {C}onformation of a system of three linked peptide units},
  journal   = {Biopolymers},
  year      = {1968},
  volume    = {6},
  number    = {10},
  pages     = {1425--1436},
  doi       = {10.1002/bip.1968.360061006}
}

@article{smith1980,
  author    = {Smith, James A. and Pease, Linda G.},
  title     = {Reverse turns in peptides and proteins},
  journal   = {CRC Critical Reviews in Biochemistry},
  year      = {1980},
  volume    = {8},
  number    = {4},
  pages     = {315--399},
  doi       = {10.3109/10409238009105479}
}

@article{chakrabartty1993,
  author    = {Chakrabartty, Avijit and Doig, Andrew J. and Baldwin, Robert L.},
  title     = {Helix capping propensities in peptides parallel those in proteins},
  journal   = {Proceedings of the National Academy of Sciences},
  year      = {1993},
  volume    = {90},
  number    = {23},
  pages     = {11332--11336},
  doi       = {10.1073/pnas.90.23.11332}
}

@article{jiang2024,
  author    = {Jiang, Huiqiang and Li, Yucheng and Zhang, Chengruidong and
               Wu, Qianhui and Luo, Xufang and Ahn, Surin and Han, Zhenhua and
               Abdi, Amir H. and Li, Dongsheng and Lin, Chin-Yew and others},
  title     = {{MInference} 1.0: Accelerating Pre-filling for Long-Context
               {LLMs} via Dynamic Sparse Attention},
  journal   = {arXiv preprint arXiv:2407.02490},
  year      = {2024}
}

@article{netsurfp,
  author    = {Klausen, Michael Schantz and Jespersen, Martin Closter and Nielsen, Henrik and Jensen, Kamilla Kjaergaard and Jurtz, Vanessa Isabell and Soenderby, Casper Kaae and Sommer, Morten Otto Alexander and Winther, Ole and Nielsen, Morten and Petersen, Bent and others},
  title     = {{N}et{S}urf{P}-2.0: Improved prediction of protein structural features by integrated deep learning},
  journal   = {Proteins: Structure, Function, and Bioinformatics},
  year      = {2019},
  publisher = {Wiley Online Library}
}

@article{casp,
  author    = {Moult, John and Fidelis, Krzysztof and Kryshtafovych, Andriy and Schwede, Torsten and Tramontano, Anna},
  title     = {Critical assessment of methods of protein structure prediction ({CASP})-{R}ound {XII}},
  journal   = {Proteins: Structure, Function, and Bioinformatics},
  year      = {2018},
  volume    = {86},
  pages     = {7--15},
  doi       = {10.1002/prot.25415},
  issn      = {08873585},
  publisher = {John Wiley {\&} Sons, Ltd},
  url       = {http://doi.wiley.com/10.1002/prot.25415}
}

@article{sarkisyan2016,
  author    = {Sarkisyan, Karen S. and Bolotin, Dmitry A. and Meer, Margarita V. and Usmanova, Dinara R. and Mishin, Alexander S. and Sharonov, George V. and Ivankov, Dmitry N. and Bozhanova, Nina G. and Baranov, Mikhail S. and Soylemez, Onuralp and others},
  title     = {Local fitness landscape of the green fluorescent protein},
  journal   = {Nature},
  volume    = {533},
  number    = {7603},
  pages     = {397},
  year      = {2016},
  publisher = {Nature Publishing Group}
}

@article{uniprot2019,
  author  = {{The UniProt Consortium}},
  title   = {UniProt: a worldwide hub of protein knowledge},
  journal = {Nucleic Acids Research},
  year    = {2019},
  volume  = {47},
  number  = {D1},
  pages   = {D506--D515},
  doi     = {10.1093/nar/gky1049}
}

@article{iupac1984,
  author  = {{IUPAC-IUB Joint Commission on Biochemical Nomenclature}},
  title   = {Nomenclature and symbolism for amino acids and peptides (Recommendations 1983)},
  journal = {Pure and Applied Chemistry},
  year    = {1984},
  volume  = {56},
  number  = {5},
  pages   = {595--624},
  doi     = {10.1351/pac198456050595}
}

@article{cuff1999,
  author  = {Cuff, James A. and Barton, Geoffrey J.},
  title   = {Evaluation and improvement of multiple sequence methods for protein secondary structure prediction},
  journal = {Proteins: Structure, Function, and Bioinformatics},
  year    = {1999},
  volume  = {34},
  number  = {4},
  pages   = {508--519},
  doi     = {10.1002/(SICI)1097-0134(19990301)34:4<508::AID-PROT10>3.0.CO;2-4}
}

@article{yang2018,
  author  = {Yang, Yuedong and Gao, Jianzhao and Wang, Jihua and Heffernan, Rhys and Hanson, Jack and Paliwal, Kuldip and Zhou, Yaoqi},
  title   = {Sixty-five years of the long march in protein secondary structure prediction: the final stretch?},
  journal = {Briefings in Bioinformatics},
  year    = {2018},
  volume  = {19},
  number  = {3},
  pages   = {482--494},
  doi     = {10.1093/bib/bbw129}
}

@article{horovitz1990strategy,
  author  = {Horovitz, Amnon and Fersht, Alan R.},
  title   = {Strategy for Analysing the Co-operativity of Intramolecular Interactions in Peptides and Proteins},
  journal = {Journal of Molecular Biology},
  volume  = {214},
  number  = {3},
  pages   = {613--617},
  year    = {1990}
}

@article{wells1990additivity,
  author  = {Wells, James A.},
  title   = {Additivity of Mutational Effects in Proteins},
  journal = {Biochemistry},
  volume  = {29},
  number  = {37},
  pages   = {8509--8517},
  year    = {1990}
}

@article{poelwijk2016context,
  author  = {Poelwijk, Frank J. and Krishna, Vijay and Ranganathan, Rama},
  title   = {The Context-Dependence of Mutations: A Linkage of Formalisms},
  journal = {PLOS Computational Biology},
  volume  = {12},
  number  = {6},
  pages   = {e1004771},
  year    = {2016},
  doi     = {10.1371/journal.pcbi.1004771}
}

@article{poelwijk2019learning,
  author  = {Poelwijk, Frank J. and Socolich, Michael and Ranganathan, Rama},
  title   = {Learning the Pattern of Epistasis Linking Genotype and Phenotype in a Protein},
  journal = {Nature Communications},
  volume  = {10},
  pages   = {4213},
  year    = {2019},
  doi     = {10.1038/s41467-019-12130-8}
}

@article{weinreich2013should,
  author  = {Weinreich, Daniel M. and Lan, Y. and Wylie, Christopher S. and Heckendorn, Robert B.},
  title   = {Should Evolutionary Geneticists Worry about Higher-Order Epistasis?},
  journal = {Current Opinion in Genetics \& Development},
  volume  = {23},
  number  = {6},
  pages   = {700--707},
  year    = {2013}
}

@article{sailer2017high,
  author  = {Sailer, Zachary R. and Harms, Michael J.},
  title   = {High-Order Epistasis Shapes Evolutionary Trajectories},
  journal = {PLOS Computational Biology},
  volume  = {13},
  number  = {5},
  pages   = {e1005541},
  year    = {2017},
  doi     = {10.1371/journal.pcbi.1005541}
}

@article{wu2016adaptation,
  author  = {Wu, Nicholas C. and Dai, Liang and Olson, Catherine A. and Lloyd-Smith, James O. and Sun, Ren},
  title   = {Adaptation in Protein Fitness Landscapes Is Facilitated by Indirect Paths},
  journal = {eLife},
  volume  = {5},
  pages   = {e16965},
  year    = {2016},
  doi     = {10.7554/eLife.16965}
}

@inproceedings{vaswani2017attention,
  author    = {Vaswani, Ashish and Shazeer, Noam and Parmar, Niki and Uszkoreit, Jakob and Jones, Llion and Gomez, Aidan N. and Kaiser, {\L}ukasz and Polosukhin, Illia},
  title     = {Attention Is All You Need},
  booktitle = {Advances in Neural Information Processing Systems},
  year      = {2017}
}

@inproceedings{clift2020logic,
  author    = {Clift, James and Murfet, Daniel},
  title     = {Logic and the 2-Simplicial Transformer},
  booktitle = {International Conference on Learning Representations},
  year      = {2020}
}

@inproceedings{sanford2024representational,
  author    = {Sanford, Clayton and Hsu, Daniel and Telgarsky, Matus},
  title     = {Representational Strengths and Limitations of Transformers},
  booktitle = {Advances in Neural Information Processing Systems},
  year      = {2024}
}

@article{alquraishi2019,
  title   = {{P}rotein{N}et: a standardized data set for machine learning of protein structure},
  author  = {AlQuraishi, Mohammed},
  journal = {BMC Bioinformatics},
  volume  = {20},
  number  = {1},
  pages   = {311},
  year    = {2019},
  doi     = {10.1186/s12859-019-2932-0}
}

@inproceedings{kozachinskiy2025strassen,
  author    = {Kozachinskiy, Alexander and Urrutia, Felipe and Jimenez, Hector and
               Steifer, Tomasz and Pizarro, Germ{\'a}n and Fuentes, Mat{\'i}as and
               Meza, Francisco and Calderon, Cristian B. and Rojas, Crist{\'o}bal},
  title     = {Strassen Attention, Split {VC} Dimension and Compositionality in Transformers},
  booktitle = {Advances in Neural Information Processing Systems},
  year      = {2025},
  note      = {arXiv:2501.19215}
}

@article{chakrabarti2026poly,
  author  = {Chakrabarti, Sayak and Pitassi, Toniann and Alman, Josh},
  title   = {Poly-attention: A General Scheme for Higher-Order Self-Attention},
  journal = {arXiv preprint arXiv:2602.02422},
  year    = {2026}
}

@inproceedings{peng2024composition,
  title     = {On Limitations of the Transformer Architecture},
  author    = {Peng, Binghui and Narayanan, Srini and Papadimitriou, Christos},
  booktitle = {Proceedings of the 1st Conference on Language Modeling},
  year      = {2024}
}

@article{hahn2020,
  title   = {Theoretical Limitations of Self-Attention in Neural Sequence Models},
  author  = {Hahn, Michael},
  journal = {Transactions of the Association for Computational Linguistics},
  volume  = {8},
  pages   = {156--171},
  year    = {2020},
  doi     = {10.1162/tacl_a_00306}
}

@inproceedings{chiang2022overcoming,
  title     = {Overcoming a Theoretical Limitation of Self-Attention},
  author    = {Chiang, David and Cholak, Peter},
  booktitle = {Proceedings of the 60th Annual Meeting of the Association for Computational Linguistics (Volume 1: Long Papers)},
  pages     = {7654--7664},
  year      = {2022},
  publisher = {Association for Computational Linguistics},
  doi       = {10.18653/v1/2022.acl-long.527}
}

@inproceedings{bhattamishra2020,
  title     = {On the Ability and Limitations of Transformers to Recognize Formal Languages},
  author    = {Bhattamishra, Satwik and Ahuja, Kabir and Goyal, Navin},
  booktitle = {Proceedings of the 2020 Conference on Empirical Methods in Natural Language Processing (EMNLP)},
  pages     = {7096--7116},
  year      = {2020},
  publisher = {Association for Computational Linguistics},
  doi       = {10.18653/v1/2020.emnlp-main.576}
}

@article{jumper2021highly,
  title   = {Highly accurate protein structure prediction with AlphaFold},
  author  = {Jumper, John and Evans, Richard and Pritzel, Alexander and Green, Tim
             and Figurnov, Michael and Ronneberger, Olaf and Tunyasuvunakool, Kathryn
             and Bates, Russ and {\v{Z}}{\'i}dek, Augustin and Potapenko, Anna
             and Bridgland, Alex and Meyer, Clemens and Kohl, Simon A. A.
             and Ballard, Andrew J. and Cowie, Andrew and Romera-Paredes, Bernardino
             and Nikolov, Stanislav and Jain, Rishub and Adler, Jonas and Back, Trevor
             and Petersen, Stig and Reiman, David and Clancy, Ellen and Zielinski, Michal
             and Steinegger, Martin and Pacholska, Michalina and Berghammer, Tamas
             and Bodenstein, Sebastian and Silver, David and Vinyals, Oriol
             and Senior, Andrew W. and Kavukcuoglu, Koray and Kohli, Pushmeet
             and Hassabis, Demis},
  journal = {Nature},
  volume  = {596},
  number  = {7873},
  pages   = {583--589},
  year    = {2021},
  doi     = {10.1038/s41586-021-03819-2}
}

@article{morcos2011direct,
  title   = {Direct-coupling analysis of residue coevolution captures native contacts across many protein families},
  author  = {Morcos, Faruck and Pagnani, Andrea and Lunt, Bryan and Bertolino, Arianna
             and Marks, Debora S. and Sander, Chris and Zecchina, Riccardo
             and Onuchic, Jos{\'e} N. and Hwa, Terence and Weigt, Martin},
  journal = {Proceedings of the National Academy of Sciences},
  volume  = {108},
  number  = {49},
  pages   = {E1293--E1301},
  year    = {2011},
  doi     = {10.1073/pnas.1111471108}
}

@article{bai2021hypergraph,
  author  = {Bai, Song and Zhang, Feihu and Torr, Philip H. S.},
  title   = {Hypergraph Convolution and Hypergraph Attention},
  journal = {Pattern Recognition},
  volume  = {110},
  pages   = {107637},
  year    = {2021},
  publisher = {Elsevier}
}

@article{beltagy2020longformer,
  author  = {Beltagy, Iz and Peters, Matthew E. and Cohan, Arman},
  title   = {Longformer: The Long-Document Transformer},
  journal = {arXiv preprint arXiv:2004.05150},
  year    = {2020}
}

@inproceedings{zaheer2020bigbird,
  author    = {Zaheer, Manzil and Guruganesh, Guru and Dubey, Kumar Avinava and Ainslie, Joshua and Alberti, Chris and Ontanon, Santiago and Pham, Philip and Ravula, Anirudh and Wang, Qifan and Yang, Li and others},
  title     = {{B}ig {B}ird: Transformers for Longer Sequences},
  booktitle = {Advances in Neural Information Processing Systems},
  year      = {2020}
}

@inproceedings{dao2022flashattention,
  author    = {Dao, Tri and Fu, Dan and Ermon, Stefano and Rudra, Atri and R{\'e}, Christopher},
  title     = {{F}lash{A}ttention: Fast and Memory-Efficient Exact Attention with {IO}-{A}wareness},
  booktitle = {Advances in Neural Information Processing Systems},
  volume    = {35},
  pages     = {16344--16359},
  year      = {2022}
}

@inproceedings{shen2018biblostan,
  title={Bi-Directional Block Self-Attention for Fast and Memory-Efficient Sequence Modeling},
  author={Shen, Tao and Zhou, Tianyi and Long, Guodong and Jiang, Jing and Zhang, Chengqi},
  booktitle={International Conference on Learning Representations (ICLR)},
  year={2018},
  url={https://arxiv.org/abs/1804.00857}
}

@inproceedings{qiu2020blockbert,
  title={Blockwise Self-Attention for Long Document Understanding},
  author={Qiu, Jiezhong and Ma, Hao and Levy, Omer and Yih, Wen-tau and Wang, Sinong and Tang, Jie},
  booktitle={Findings of the Association for Computational Linguistics},
  year={2020},
  url={https://arxiv.org/abs/1911.02972}
}
\bibliographystyle{iclr2027_conference}

\clearpage
\appendix
\numberwithin{table}{section}
\numberwithin{figure}{section}
%

\section*{Appendix}
We provide experimental details, full results, and reproducibility information
omitted from the main paper. The appendices are organized as follows:
\begin{itemize}
    \item Appendix~\ref{app:hparams}: complete hyperparameter specification.
    \item Appendix~\ref{app:protocol}: diagnostic-task protocol and conventions.
    \item Appendix~\ref{app:parity}: additional \textsc{parity} results.
    \item Appendix~\ref{app:match}: \textsc{match2} and \textsc{match3} setup and results.
    \item Appendix~\ref{app:tape}: full TAPE capacity results.
    \item Appendix~\ref{app:memory}: compute and peak-memory measurements.
    \item Appendix~\ref{app:coverage}: triadic-window coverage and depth.
    \item Appendix~\ref{app:code}: code package and reproducibility details.

\end{itemize}

\section{Complete hyperparameter specification}
\label{app:hparams}

Tables~\ref{tab:app_hp_data}--\ref{tab:app_hp_optim} give the complete experimental
configuration. Within each task's capacity sweep, only $d_{\mathrm{model}}$
varies.

\begin{table}[!h]
\centering
\scriptsize
\setlength{\tabcolsep}{3.5pt}
\renewcommand{\arraystretch}{1.12}
\caption{\textbf{Data and task specification.} ``---'' denotes a non-applicable
setting. The \textsc{match} modulus $M$ is calibrated by sequence length and
task order (Section~\ref{app:match_setup}).}
\label{tab:app_hp_data}
\begin{tabular}{@{}l ccccc@{}}
\toprule
& \textbf{\textsc{parity}/} & \textbf{\textsc{match2}/} & \textbf{\shortstack{Secondary\\Structure}}
& \textbf{\shortstack{Contact\\Prediction}} & \textbf{Fluorescence} \\
& \textbf{\textsc{majority}} & \textbf{\textsc{match3}} & & & \\
\midrule
Train / valid / test
  & 3{,}000 / --- / 800 & 30{,}000 / --- / 2{,}000
  & 8{,}678 / 2{,}170 / 3 sets
  & 25{,}299 / 224 / 40 & 21{,}446 / 5{,}362 / 27{,}217 \\
Sequence length & $L{=}16$ & $N\in\{6,8\}$ & pad to 512 & crop to 256 & fixed at 237 \\
Vocabulary & 2 (binary) & $M$ (calibrated) & 30 (IUPAC) & 30 (IUPAC) & 30 (IUPAC) \\
Label & per position & per position & per residue & per pair & per sequence \\
Test split & held out & held out & CB513 / CASP12 / TS115 & CASP12 & $H_d\!\geq\!4$ \\
\bottomrule
\end{tabular}
\end{table}

\begin{table}[!h]
\centering
\scriptsize
\setlength{\tabcolsep}{4pt}
\renewcommand{\arraystretch}{1.12}
\caption{\textbf{Architecture and attention.} Only the attention mechanism
varies within a task; every other quantity here is shared across mechanisms at a
given width.}
\label{tab:app_hp_arch}
\begin{tabular}{@{}l ccccc@{}}
\toprule
& \textbf{\textsc{parity}/} & \textbf{\textsc{match2}/} & \textbf{\shortstack{Secondary\\Structure}}
& \textbf{\shortstack{Contact\\Prediction}} & \textbf{Fluorescence} \\
& \textbf{\textsc{majority}} & \textbf{\textsc{match3}} & & & \\
\midrule
$d_{\mathrm{model}}$ range & 8--128 & 8--256 & 32--512 & 32--256 & 32--256 \\
Layers & \shortstack{1, 6, 12 (grid); 1 (capacity);\\1, 2, 4 (coverage)} & 1 & 12 & 12 & 12 \\
Heads & 4 & 4 & 8 & 8 & 8 \\
FFN width & none & none & $2 d_{\mathrm{model}}$ & $2 d_{\mathrm{model}}$
  & $2 d_{\mathrm{model}}$ \\
Dropout & 0 & 0 & 0.4 & 0.1 & 0.1 \\
Task head & linear, 2-way & linear, 2-way & per-residue cls.
  & pairwise, symmetric & mean-pool reg. \\
\midrule
Block $\ell$ / stride $s$ & single block & $\ell{=}s{=}N$ & 30 / 15 & 32 / 16 & 30 / 15 \\
Triadic window $w$ & 7 (uniform across $k$) & $2N{-}1$ & 5 & 5 & 5 \\
Low-rank $U$ rank $r$ & 8 & 8 & 8 & 8 & 8 \\
\bottomrule
\end{tabular}
\end{table}

\begin{table}[!h]
\centering
\scriptsize
\setlength{\tabcolsep}{4pt}
\renewcommand{\arraystretch}{1.12}
\caption{\textbf{Optimization, loss, and evaluation settings.}}
\label{tab:app_hp_optim}
\begin{tabular}{@{}l ccccc@{}}
\toprule
& \textbf{\textsc{parity}/} & \textbf{\textsc{match2}/} & \textbf{\shortstack{Secondary\\Structure}}
& \textbf{\shortstack{Contact\\Prediction}} & \textbf{Fluorescence} \\
& \textbf{\textsc{majority}} & \textbf{\textsc{match3}} & & & \\
\midrule
Optimizer & Adam & Adam & Adam & AdamW & Adam \\
Learning rate & $2\times10^{-3}$ & $3\times10^{-3}$ & $10^{-4}$ & $10^{-4}$
  & $10^{-4}$ \\
Weight decay & 0 & 0 & 0 & 0.01 & 0 \\
LR schedule & none & none & none & OneCycle (10\% warm-up)
  & cosine (6\% warm-up) \\
Gradient clipping & none & none & none & 1.0 & 1.0 \\
Batch size & 128 & 256 & 16 & 8 & 16 \\
Epochs & 40 & 40 & 10 & 8 & 10 \\
\midrule
Loss & CE, ignore $-100$ & CE & token CE, ignore $-100$
  & weighted CE, ignore $-1$ & MSE \\
Class weighting & --- & --- & --- & $5\times$ positive & --- \\
\midrule
Metric & accuracy & accuracy & Q3 accuracy & P@$L/5$ long-range
  & Spearman $\rho$ \\
Checkpoint & last epoch & last epoch & best val.\ loss & last epoch
  & best val.\ $\rho$ \\
Seeds & 0, 1, 2 & 0, 1, 2 & 42, 456, 608 & 0, 1, 2 & 42, 456, 608 \\
\bottomrule
\end{tabular}
\end{table}

\paragraph{Losses, stated per task.}
Ignore index $-100$ masks unlabelled or padded token positions, and $-1$ masks
unscored residue pairs in Contact Prediction. Contact Prediction upweights
positives by $5\times$ because contacts comprise approximately $2\%$ of scored
pairs.

\paragraph{Checkpoint selection on the TAPE tasks.}
TAPE results use the task-specific validation criterion in
Table~\ref{tab:app_hp_optim}, except Contact Prediction, which reports the final
epoch; test data are never used for selection. Diagnostic results instead use the
final epoch of a fixed budget with no checkpoint selection
(Section~\ref{app:protocol}).

\paragraph{Notes on individual settings.}
For the diagnostic sweeps, \textsc{Pairwise-2D} and \textsc{Blockwise-2D} are identical because
the length-16 sequence forms one block; their per-seed accuracies agree, so the
tables show one pairwise row. A single triadic window, determined by the widest
offset pattern, is held fixed across $k$; Section~\ref{app:coverage} varies this
window separately.

For Contact Prediction, cropping is a random window during training and the
centred window at evaluation, so validation and test are deterministic. Proteins
shorter than 32 residues are dropped, as they cannot hold a scoreable
$|i-j|\ge6$ pair.

\paragraph{The \textsc{match} tasks.}
The frozen Fourier embedding is followed by a learned linear projection, and
positional embeddings are disabled. Runs use independent train and test streams
for each seed, no validation split, and the final epoch of the 40-epoch budget.
Tested widths are 16--128 for \textsc{match2} and 8--256 for \textsc{match3};
majority-class accuracy is 0.500--0.540.

\paragraph{Preprocessing and sequence handling.}
All protein datasets use residue-level tokenization with the TAPE IUPAC
tokenizer, a 30-token vocabulary of five control symbols and a 25-character
residue alphabet: the 20 standard amino acids, selenocysteine and pyrrolysine,
two ambiguity codes, and one unknown-residue symbol
\citep{iupac1984,uniprot2019}. Secondary Structure sequences are padded to 512,
Contact Prediction sequences are cropped to 256, and Fluorescence inputs are
fixed at 237 tokens. Padded positions are excluded from attention and evaluation
by masking.

\paragraph{A note on the epoch budget.}
Epochs are equal in number but not in optimization steps across the TAPE tasks.
At batch size 16, one epoch is $543$ steps for Secondary Structure and
$1{,}341$ for Fluorescence. Comparisons within a task are unaffected; comparisons of
convergence across tasks are not meaningful and should not be drawn.

\section{Protocol and conventions for the diagnostic tasks}
\label{app:protocol}

Diagnostic results use last-epoch accuracy over labelled test positions; chance
is $0.50$. Selecting the best epoch instead moves no cell by more than $0.010$;
the two cells that move by more than $0.005$, \textsc{Pairwise-2D} at $k=4$ and
$5$, are at chance under either convention.

Epochs-to-threshold is the first test epoch above the specified accuracy. Means
include only successful seeds and show the success count, so values based on
fewer than three seeds are conditional on success. ``$>$40'' means that no seed
crossed within the training budget.

\paragraph{Two conventions that must not be mixed.}
Tables report the mean crossing epoch among successful seeds, whereas
Figure~\ref{fig:parity_convergence} assigns non-crossing seeds epoch 41 and averages all seeds.
Consequently, figure annotations and table entries for the same configuration
may differ.

\section{\textsc{parity} results}
\label{app:parity}

\begin{table}[!h]
\centering
\footnotesize
\setlength{\tabcolsep}{4pt}
\caption{\textbf{Lower-order \textsc{parity}, the \textsc{majority} control,
and convergence.} \textbf{(a)} Mean final test accuracy $\pm$ standard deviation
for \textsc{parity}-$k$, $k\in\{1,2\}$. \textbf{(b)} Mean final accuracy for
\textsc{majority}-$k$; all standard deviations are below $0.001$.
\textbf{(c)} Mean epochs to $0.90$ among successful seeds for
\textsc{parity}-$k$ at $d_{\mathrm{model}}{=}64$; counts are shown when fewer
than three seeds cross, ``$>$40'' denotes no crossing, and \textit{n/a} denotes
an unavailable learning curve (Section~\ref{app:protocol}).}
\label{tab:app_parity_extra}
\textbf{(a)} \textsc{parity}-$k$, $k{=}1,2$\\[2pt]
\begin{tabular}{@{}ll ccc G ccc@{}}
\toprule
& & \multicolumn{3}{c}{$k{=}1$} & \multicolumn{3}{c}{$k{=}2$} \\
\cmidrule(lr){3-5}\cmidrule(lr){6-8}
$d$ & mechanism & $D{=}1$ & $6$ & $12$ & $D{=}1$ & $6$ & $12$ \\
\midrule
$\mathbf{32}$ & \textsc{Pairwise-2D} & 1.000\sd{.000} & 1.000\sd{.000} & 1.000\sd{.000} & 1.000\sd{.000} & 1.000\sd{.000} & 1.000\sd{.000} \\
& \textsc{Blockwise-3D} & 1.000\sd{.000} & 1.000\sd{.000} & 1.000\sd{.000} & 1.000\sd{.000} & 1.000\sd{.000} & 1.000\sd{.000} \\
& \textsc{HOMA-add} & 1.000\sd{.000} & 1.000\sd{.000} & 1.000\sd{.000} & 1.000\sd{.000} & 1.000\sd{.000} & 1.000\sd{.000} \\
& \textsc{HOMA} & 1.000\sd{.000} & 1.000\sd{.000} & 1.000\sd{.000} & 1.000\sd{.000} & 1.000\sd{.000} & 1.000\sd{.000} \\
\addlinespace[2pt]\midrule
$\mathbf{64}$ & \textsc{Pairwise-2D} & 1.000\sd{.000} & 1.000\sd{.000} & 1.000\sd{.000} & 0.983\sd{.030} & 1.000\sd{.000} & 1.000\sd{.000} \\
& \textsc{Blockwise-3D} & 1.000\sd{.000} & 1.000\sd{.000} & 1.000\sd{.000} & 1.000\sd{.000} & 1.000\sd{.000} & 1.000\sd{.000} \\
& \textsc{HOMA-add} & 1.000\sd{.000} & 1.000\sd{.000} & 1.000\sd{.000} & 1.000\sd{.000} & 1.000\sd{.000} & 1.000\sd{.000} \\
& \textsc{HOMA} & 1.000\sd{.000} & 1.000\sd{.000} & 1.000\sd{.000} & 1.000\sd{.000} & 1.000\sd{.000} & 1.000\sd{.000} \\
\bottomrule
\end{tabular}

\vspace{8pt}
\textbf{(b)} \textsc{majority}-$k$ control across network depth $D$ and embedding size $d$\\[2pt]
{\scriptsize\setlength{\tabcolsep}{2.4pt}
\begin{tabular}{@{}ll ccc G ccc G ccc G ccc G ccc@{}}
\toprule
& & \multicolumn{3}{c}{$k{=}1$} & \multicolumn{3}{c}{$k{=}2$} & \multicolumn{3}{c}{$k{=}3$} & \multicolumn{3}{c}{$k{=}4$} & \multicolumn{3}{c}{$k{=}5$} \\
\cmidrule(lr){3-5}\cmidrule(lr){6-8}\cmidrule(lr){9-11}\cmidrule(lr){12-14}\cmidrule(lr){15-17}
$d$ & mechanism & $D{=}1$ & $6$ & $12$ & $D{=}1$ & $6$ & $12$ & $D{=}1$ & $6$ & $12$ & $D{=}1$ & $6$ & $12$ & $D{=}1$ & $6$ & $12$ \\
\midrule
$\mathbf{32}$ & \textsc{Pairwise-2D} & 1.000 & 1.000 & 1.000 & 1.000 & 1.000 & 1.000 & 1.000 & 1.000 & 1.000 & 1.000 & 1.000 & 1.000 & 1.000 & 1.000 & 1.000 \\
& \textsc{Blockwise-3D} & 1.000 & 1.000 & 1.000 & 1.000 & 1.000 & 1.000 & 1.000 & 1.000 & 1.000 & 1.000 & 1.000 & 1.000 & 1.000 & 1.000 & 1.000 \\
& \textsc{HOMA-add} & 1.000 & 1.000 & 1.000 & 1.000 & 1.000 & 1.000 & 1.000 & 1.000 & 1.000 & 1.000 & 1.000 & 1.000 & 1.000 & 1.000 & 1.000 \\
& \textsc{HOMA} & 1.000 & 1.000 & 1.000 & 1.000 & 1.000 & 1.000 & 1.000 & 1.000 & 1.000 & 1.000 & 1.000 & 1.000 & 1.000 & 1.000 & 1.000 \\
\addlinespace[2pt]\midrule
$\mathbf{64}$ & \textsc{Pairwise-2D} & 1.000 & 1.000 & 1.000 & 1.000 & 1.000 & 1.000 & 1.000 & 1.000 & 1.000 & 1.000 & 1.000 & 1.000 & 1.000 & 1.000 & 1.000 \\
& \textsc{Blockwise-3D} & 1.000 & 1.000 & 1.000 & 1.000 & 1.000 & 1.000 & 1.000 & 1.000 & 1.000 & 1.000 & 1.000 & 1.000 & 1.000 & 1.000 & 1.000 \\
& \textsc{HOMA-add} & 1.000 & 1.000 & 1.000 & 1.000 & 1.000 & 1.000 & 1.000 & 1.000 & 1.000 & 1.000 & 1.000 & 1.000 & 1.000 & 1.000 & 1.000 \\
& \textsc{HOMA} & 1.000 & 1.000 & 1.000 & 1.000 & 1.000 & 1.000 & 1.000 & 1.000 & 1.000 & 1.000 & 1.000 & 1.000 & 1.000 & 1.000 & 1.000 \\
\bottomrule
\end{tabular}}

\vspace{8pt}
\textbf{(c)} Epochs to $0.90$ on \textsc{parity}-$k$, $d_{\mathrm{model}}{=}64$\\[2pt]
\begin{tabular}{@{}l ccc G ccc G ccc@{}}
\toprule
& \multicolumn{3}{c}{$k{=}3$} & \multicolumn{3}{c}{$k{=}4$} & \multicolumn{3}{c}{$k{=}5$} \\
\cmidrule(lr){2-4}\cmidrule(lr){5-7}\cmidrule(lr){8-10}
mechanism & $D{=}1$ & $6$ & $12$ & $D{=}1$ & $6$ & $12$ & $D{=}1$ & $6$ & $12$ \\
\midrule
\textsc{Pairwise-2D} & \textit{n/a} & 10.3 & 10.0 & $>$40 & $>$40 & $>$40 & $>$40 & $>$40 & $>$40 \\
\textsc{Blockwise-3D} & \textit{n/a} & 3.0 & 2.7 & 12.3 & 6.7 & 6.0 & 20.0 & 9.7 & 6.0 \\
\textsc{HOMA-add} & 4.3 & 4.0 & 5.0 & 12.7 & 10.3 & 10.7 & 24.3 & 11.3 & 9.7 \\
\textsc{HOMA} & \textit{n/a} & 3.0 & 3.0 & 5.3 & 4.3 & 4.0 & 7.7 & 5.7 & 5.7 \\
\bottomrule
\end{tabular}
\end{table}

\begin{table}[!h]
\centering
\footnotesize
\setlength{\tabcolsep}{4pt}
\caption{\textbf{Capacity on \textsc{parity}-3.} Trainable parameters, mean
final test accuracy over three seeds, and mean epochs to $0.90$ among successful
seeds. One attention layer without a residual connection; a separate run series
from the depth grid of Table~\ref{tab:app_parity_extra}.}
\label{tab:app_parity3_capacity}
\begin{tabular}{@{}r rcc rcc rcc@{}}
\toprule
& \multicolumn{3}{c}{\textsc{Pairwise-2D}} & \multicolumn{3}{c}{\textsc{Blockwise-3D}} & \multicolumn{3}{c}{\textsc{HOMA}} \\
\cmidrule(lr){2-4}\cmidrule(lr){5-7}\cmidrule(l){8-10}
$d_{\mathrm{model}}$ & params & acc & ep@0.90 & params & acc & ep@0.90 & params & acc & ep@0.90 \\
\midrule
$8$ & 466 & 0.586\sd{.041} & $>$40 & 594 & 0.946\sd{.058} & 22.5\,{\tiny(2/3)} & 1{,}492 & 0.982\sd{.030} & 19.3 \\
$16$ & 1{,}442 & 0.879\sd{.082} & 35.5\,{\tiny(2/3)} & 1{,}698 & 0.989\sd{.019} & 10.0 & 3{,}366 & 1.000\sd{.000} & 6.0 \\
$32$ & 4{,}930 & 0.978\sd{.037} & 21.0 & 5{,}442 & 1.000\sd{.000} & 4.7 & 8{,}650 & 1.000\sd{.000} & 5.0 \\
$64$ & 18{,}050 & 1.000\sd{.000} & 11.7 & 19{,}074 & 1.000\sd{.000} & 3.0 & 25{,}362 & 1.000\sd{.000} & 3.0 \\
\bottomrule
\end{tabular}
\end{table}

\FloatBarrier
\section{\textsc{match2} and \textsc{match3}: setup and results}
\label{app:match}

\subsection{Task setup, and where $M$ and $N$ stop}
\label{app:match_setup}

\paragraph{Task.}
The task definitions are given in Section~\ref{sec:setup}. Here $j$ and $z$
range over the full sequence and may equal each other or $i$. Each token is
encoded by a frozen sine--cosine Fourier feature followed by a learned
projection; $d_{\mathrm{model}}$ therefore limits the represented frequencies.

\paragraph{Choice of the modulus $M$.}
For approximately uniform tokens, $\Pr(y_i{=}0)\approx e^{-L_N/M}$, so balanced
labels require $M\approx L_N/\ln 2$, where $L_N$ is the number of candidate
partners per query: $L_N=N$ for \textsc{match2} and
$L_N=\binom{N+1}{2}$ for \textsc{match3}. Hence $M=\Theta(N)$ for \textsc{match2} and
$M=\Theta(N^2)$ for \textsc{match3}. For each $(N,\mathrm{order})$, we select
the grid value of $M$ whose measured positive rate is closest to $0.5$; all
reported rates lie within $0.04$ of $0.50$.

\paragraph{Why the reported lengths stop at $N{=}8$.}
A complete real Fourier basis requires approximately $M-1$ coordinates. Because
\textsc{match3} uses $M=\Theta(N^2)$, its required embedding width also grows
quadratically with $N$. We therefore stop at $N=8$: the tested widths include a
complete basis for $N=6$ ($M=30$) and $N=8$ ($M=56$), whereas longer sequences
would confound attention order with incomplete input features.

\subsection{Results}

\begin{table}[!h]
\centering
\footnotesize
\setlength{\tabcolsep}{4pt}
\caption{\textbf{\textsc{match2} control.} Mean final test accuracy $\pm$
standard deviation over three seeds across sequence lengths $N$, calibrated
moduli $M$, and embedding widths $d_{\mathrm{model}}$.}
\label{tab:app_match2}
\begin{tabular}{@{}ll cccc@{}}
\toprule
& & \multicolumn{4}{c}{test accuracy at $d_{\mathrm{model}}$} \\
\cmidrule(l){3-6}
$N\,(M)$ & mechanism & $16$ & $32$ & $64$ & $128$ \\
\midrule
$6\,(9)$ & \textsc{Pairwise-2D} & 1.000\sd{.000} & 1.000\sd{.000} & 1.000\sd{.000} & 1.000\sd{.000} \\
 & \textsc{Blockwise-3D} & 1.000\sd{.000} & 1.000\sd{.000} & 1.000\sd{.000} & 1.000\sd{.000} \\
 & \textsc{HOMA-add} & 1.000\sd{.000} & 1.000\sd{.000} & 1.000\sd{.000} & 1.000\sd{.000} \\
 & \textsc{HOMA} & 1.000\sd{.000} & 1.000\sd{.000} & 1.000\sd{.000} & 1.000\sd{.000} \\
\addlinespace[3pt]
$8\,(14)$ & \textsc{Pairwise-2D} & 1.000\sd{.000} & 1.000\sd{.000} & 1.000\sd{.000} & 0.979\sd{.036} \\
 & \textsc{Blockwise-3D} & 1.000\sd{.000} & 1.000\sd{.000} & 1.000\sd{.000} & 1.000\sd{.000} \\
 & \textsc{HOMA-add} & 1.000\sd{.000} & 1.000\sd{.000} & 1.000\sd{.000} & 1.000\sd{.000} \\
 & \textsc{HOMA} & 1.000\sd{.000} & 1.000\sd{.000} & 1.000\sd{.000} & 1.000\sd{.000} \\
\bottomrule
\end{tabular}
\end{table}

\begin{table}[!h]
\centering
\scriptsize
\setlength{\tabcolsep}{3pt}
\caption{\textbf{Extended \textsc{match3} width sweep.} Mean final test
accuracy $\pm$ standard deviation over three seeds. Widths 16--128 reproduce
Table~\ref{tab:match3} for completeness; widths 8 and 256 are additional.}
\label{tab:app_match3}
\begin{tabular}{@{}ll cccccc@{}}
\toprule
& & \multicolumn{6}{c}{test accuracy at $d_{\mathrm{model}}$} \\
\cmidrule(l){3-8}
$N\,(M)$ & mechanism & $8$ & $16$ & $32$ & $64$ & $128$ & $256$ \\
\midrule
$6\,(30)$ & \textsc{Pairwise-2D} & 0.760\sd{.008} & 0.816\sd{.009} & 0.835\sd{.001} & 0.830\sd{.003} & 0.783\sd{.009} & 0.776\sd{.026} \\
 & \textsc{Blockwise-3D} & 0.770\sd{.015} & 0.836\sd{.007} & 0.921\sd{.012} & 0.958\sd{.008} & 0.927\sd{.028} & 0.774\sd{.008} \\
 & \textsc{HOMA-add} & 0.741\sd{.038} & 0.835\sd{.003} & 0.929\sd{.002} & 0.990\sd{.002} & 0.975\sd{.006} & 0.817\sd{.010} \\
 & \textsc{HOMA} & 0.786\sd{.009} & 0.857\sd{.006} & 0.967\sd{.007} & 0.993\sd{.002} & 0.991\sd{.003} & 0.819\sd{.015} \\
\addlinespace[3pt]
$8\,(56)$ & \textsc{Pairwise-2D} & 0.625\sd{.014} & 0.708\sd{.005} & 0.744\sd{.004} & 0.761\sd{.009} & 0.731\sd{.012} & 0.726\sd{.007} \\
 & \textsc{Blockwise-3D} & 0.618\sd{.012} & 0.726\sd{.013} & 0.768\sd{.005} & 0.856\sd{.004} & 0.912\sd{.004} & 0.717\sd{.018} \\
 & \textsc{HOMA-add} & 0.614\sd{.015} & 0.734\sd{.014} & 0.776\sd{.004} & 0.848\sd{.003} & 0.921\sd{.004} & 0.846\sd{.089} \\
 & \textsc{HOMA} & 0.629\sd{.022} & 0.745\sd{.009} & 0.799\sd{.011} & 0.853\sd{.003} & 0.925\sd{.008} & 0.722\sd{.019} \\
\bottomrule
\end{tabular}
\end{table}

\section{TAPE capacity study: full results}
\label{app:tape}

\begin{table}[H]
\centering
\footnotesize
\setlength{\tabcolsep}{3pt}
\caption{\textbf{TAPE results by task and width.} Mean $\pm$ standard deviation
over three seeds; the best mechanism at each width is bold. Secondary Structure
reports Q3 accuracy on CB513, CASP12, and TS115; Figure~\ref{fig:tape_grid} plots CB513.
Contact Prediction reports long-range P@$L/5$, and Fluorescence reports Spearman
$\rho$. ``--'' denotes an untested width.}
\label{tab:app_tape_tests}
\begin{tabular}{@{}ll ccccc@{}}
\toprule
& & \multicolumn{5}{c}{score at $d_{\mathrm{model}}$} \\
\cmidrule(l){3-7}
task & mechanism & $32$ & $64$ & $128$ & $256$ & $512$ \\
\midrule
\shortstack[l]{Secondary Structure\\(CB513)} & \textsc{Pairwise-2D} & 0.513\sd{.001} & 0.518\sd{.001} & 0.521\sd{.001} & 0.518\sd{.002} & 0.527\sd{.001} \\
 & \textsc{Blockwise-2D} & 0.536\sd{.001} & 0.542\sd{.002} & 0.552\sd{.004} & 0.622\sd{.004} & 0.642\sd{.005} \\
 & \textsc{Blockwise-3D} & 0.620\sd{.001} & 0.636\sd{.002} & 0.643\sd{.002} & 0.642\sd{.005} & 0.644\sd{.010} \\
 & \textsc{HOMA} & \textbf{0.625}\sd{.003} & \textbf{0.642}\sd{.001} & \textbf{0.651}\sd{.001} & \textbf{0.656}\sd{.001} & \textbf{0.659}\sd{.003} \\
\addlinespace[3pt]
\shortstack[l]{Secondary Structure\\(CASP12)} & \textsc{Pairwise-2D} & 0.542\sd{.010} & 0.543\sd{.005} & 0.539\sd{.003} & 0.544\sd{.006} & 0.538\sd{.011} \\
 & \textsc{Blockwise-2D} & 0.570\sd{.006} & 0.586\sd{.005} & 0.588\sd{.004} & 0.639\sd{.006} & 0.658\sd{.003} \\
 & \textsc{Blockwise-3D} & 0.625\sd{.007} & 0.635\sd{.006} & 0.648\sd{.006} & 0.653\sd{.006} & 0.651\sd{.009} \\
 & \textsc{HOMA} & \textbf{0.635}\sd{.007} & \textbf{0.650}\sd{.004} & \textbf{0.665}\sd{.001} & \textbf{0.664}\sd{.010} & \textbf{0.669}\sd{.007} \\
\addlinespace[3pt]
\shortstack[l]{Secondary Structure\\(TS115)} & \textsc{Pairwise-2D} & 0.557\sd{.001} & 0.564\sd{.001} & 0.563\sd{.002} & 0.563\sd{.002} & 0.569\sd{.002} \\
 & \textsc{Blockwise-2D} & 0.574\sd{.003} & 0.582\sd{.001} & 0.588\sd{.005} & 0.655\sd{.002} & 0.672\sd{.009} \\
 & \textsc{Blockwise-3D} & 0.646\sd{.005} & 0.659\sd{.003} & 0.672\sd{.003} & 0.671\sd{.006} & 0.675\sd{.008} \\
 & \textsc{HOMA} & \textbf{0.653}\sd{.004} & \textbf{0.669}\sd{.002} & \textbf{0.679}\sd{.005} & \textbf{0.685}\sd{.002} & \textbf{0.691}\sd{.002} \\
\addlinespace[3pt]
\shortstack[l]{Contact Prediction\\(P@$L/5$)} & \textsc{Pairwise-2D} & 0.040\sd{.005} & 0.060\sd{.004} & 0.059\sd{.003} & 0.088\sd{.012} & -- \\
 & \textsc{Blockwise-2D} & 0.052\sd{.002} & 0.055\sd{.006} & 0.078\sd{.009} & 0.117\sd{.005} & -- \\
 & \textsc{Blockwise-3D} & \textbf{0.095}\sd{.004} & \textbf{0.121}\sd{.006} & 0.129\sd{.002} & 0.125\sd{.008} & -- \\
 & \textsc{HOMA} & 0.089\sd{.009} & 0.106\sd{.007} & \textbf{0.134}\sd{.006} & \textbf{0.134}\sd{.001} & -- \\
\addlinespace[3pt]
Fluorescence ($\rho$) & \textsc{Pairwise-2D} & 0.625\sd{.001} & 0.637\sd{.002} & 0.652\sd{.004} & 0.655\sd{.001} & -- \\
 & \textsc{Blockwise-2D} & 0.630\sd{.001} & 0.640\sd{.003} & 0.655\sd{.001} & 0.656\sd{.001} & -- \\
 & \textsc{Blockwise-3D} & \textbf{0.642}\sd{.001} & \textbf{0.654}\sd{.001} & \textbf{0.658}\sd{.002} & \textbf{0.658}\sd{.006} & -- \\
 & \textsc{HOMA} & 0.633\sd{.007} & 0.649\sd{.004} & 0.658\sd{.001} & 0.657\sd{.004} & -- \\
\bottomrule
\end{tabular}
\end{table}

\section{Compute and peak memory}
\label{app:memory}

Table~\ref{tab:app_cost_all} reports trainable parameters, throughput, and peak
allocated GPU memory for all tested configurations.

\paragraph{The overhead at fixed width is real.}
At $d_{\mathrm{model}}{=}512$ on Secondary Structure, \textsc{HOMA} processes $31.5$k
residues/s and uses $14.80$~GB, compared with $75.9$k residues/s and $3.55$~GB
for \textsc{Blockwise-2D}. Parameter counts differ by less than $2\%$ (25.9M versus
25.5M), so the principal fixed-width overhead is compute and memory.

\begin{table}[!h]
\centering
\footnotesize
\setlength{\tabcolsep}{3pt}
\caption{\textbf{Parameters, throughput, and peak memory by task and width.}
\textit{par}: trainable parameters; \textit{tput}: thousands of residues per
second; \textit{mem}: peak allocated GB. Values are means over three seeds.
Throughput is compute-only for Secondary Structure and Fluorescence and
wall-clock-derived for Contact Prediction, so comparisons are valid within, but
not across, tasks. ``--'' denotes an untested width.}
\label{tab:app_cost_all}
\begin{tabular}{@{}r rrr rrr rrr rrr@{}}
\toprule
& \multicolumn{3}{c}{\textsc{Pairwise-2D}} & \multicolumn{3}{c}{\textsc{Blockwise-2D}} & \multicolumn{3}{c}{\textsc{Blockwise-3D}} & \multicolumn{3}{c}{\textsc{HOMA}} \\
\cmidrule(lr){2-4}\cmidrule(lr){5-7}\cmidrule(lr){8-10}\cmidrule(lr){11-13}
$d_{\mathrm{model}}$ & par & tput & mem & par & tput & mem & par & tput & mem & par & tput & mem \\
\midrule
\multicolumn{13}{@{}l}{\textit{Secondary Structure}} \\
$32$ & 120k & 165 & 2.04 & 120k & 217 & 0.42 & 126k & 138 & 0.91 & 146k & 102 & 2.14 \\
$64$ & 437k & 159 & 2.19 & 437k & 215 & 0.60 & 449k & 129 & 1.64 & 487k & 97.1 & 2.97 \\
$128$ & 1.7M & 147 & 2.50 & 1.7M & 216 & 0.99 & 1.7M & 112 & 3.11 & 1.8M & 84.3 & 4.63 \\
$256$ & 6.5M & 101 & 3.14 & 6.5M & 137 & 1.81 & 6.5M & 72.0 & 6.08 & 6.7M & 57.1 & 8.02 \\
$512$ & 25.5M & 68.2 & 4.53 & 25.5M & 75.9 & 3.55 & 25.6M & 37.8 & 12.05 & 25.9M & 31.5 & 14.80 \\
\addlinespace[3pt]
\multicolumn{13}{@{}l}{\textit{Contact Prediction}} \\
$32$ & 235k & 64.5 & 0.56 & 235k & 59.9 & 0.42 & 241k & 46.7 & 0.56 & 261k & 40.0 & 0.88 \\
$64$ & 666k & 65.6 & 0.87 & 666k & 57.5 & 0.75 & 679k & 46.5 & 1.04 & 717k & 39.2 & 1.40 \\
$128$ & 2.1M & 70.9 & 1.52 & 2.1M & 63.4 & 1.45 & 2.1M & 47.0 & 2.01 & 2.2M & 36.6 & 2.45 \\
$256$ & 7.4M & 53.3 & 2.84 & 7.4M & 51.3 & 2.79 & 7.4M & 36.9 & 3.95 & 7.6M & 32.8 & 4.48 \\
\addlinespace[3pt]
\multicolumn{13}{@{}l}{\textit{Fluorescence}} \\
$32$ & 612k & 101 & 0.91 & 612k & 79.4 & 0.60 & 618k & 54.1 & 0.81 & 638k & 42.6 & 1.34 \\
$64$ & 2.4M & 96.6 & 1.00 & 2.4M & 76.7 & 0.71 & 2.4M & 53.0 & 1.14 & 2.4M & 42.4 & 1.73 \\
$128$ & 9.5M & 96.4 & 1.24 & 9.5M & 77.5 & 0.99 & 9.5M & 52.9 & 1.88 & 9.6M & 42.1 & 2.55 \\
$256$ & 37.6M & 95.9 & 1.91 & 37.6M & 76.6 & 1.46 & 37.6M & 52.1 & 3.11 & 37.8M & 41.3 & 3.94 \\
\bottomrule
\end{tabular}
\end{table}

\FloatBarrier
\section{Triadic window coverage and depth}
\label{app:coverage}

Tables~\ref{tab:app_coverage_cliff} and~\ref{tab:app_coverage_depth} provide the
numerical results underlying Figure~\ref{fig:coverage_sweep}.

\begin{table}[!h]
\centering
\footnotesize
\setlength{\tabcolsep}{3pt}
\caption{\textbf{Triadic-window coverage on \textsc{parity}-3 at depth 1.}
Dedicated coverage-sweep runs use $d_{\mathrm{model}}=32$ and sequence length
$L=24$. The window reach is $R_w=(w-1)/2$. Mean final test accuracy $\pm$
standard deviation over three seeds. \textsc{Pairwise-2D} is independent of
$w$, so the same runs are shown for both window settings.}
\label{tab:app_coverage_cliff}
\begin{tabular}{@{}ll cccccc@{}}
\toprule
& & \multicolumn{6}{c}{interaction reach $R$} \\
\cmidrule(l){3-8}
window & mechanism & $1$ & $2$ & $3$ & $4$ & $5$ & $6$ \\
\midrule
$w{=}3$ {\tiny($R_w{=}1$)} & \textsc{Pairwise-2D} & 0.907\sd{.024} & 0.951\sd{.013} & 0.913\sd{.035} & 0.884\sd{.049} & 0.816\sd{.080} & 0.857\sd{.097} \\
 & \textsc{Blockwise-3D} & 1.000\sd{.000} & 0.501\sd{.009} & 0.496\sd{.008} & 0.499\sd{.001} & 0.500\sd{.003} & 0.501\sd{.004} \\
 & \textsc{HOMA-add} & 1.000\sd{.000} & 0.858\sd{.065} & 0.936\sd{.071} & 0.862\sd{.074} & 0.854\sd{.046} & 0.829\sd{.065} \\
 & \textsc{HOMA} & 1.000\sd{.000} & 0.964\sd{.015} & 0.968\sd{.031} & 0.915\sd{.055} & 0.909\sd{.122} & 0.986\sd{.024} \\
\addlinespace[3pt]
$w{=}7$ {\tiny($R_w{=}3$)} & \textsc{Pairwise-2D} & 0.907\sd{.024} & 0.951\sd{.013} & 0.913\sd{.035} & 0.884\sd{.049} & 0.816\sd{.080} & 0.857\sd{.097} \\
 & \textsc{Blockwise-3D} & 1.000\sd{.000} & 1.000\sd{.000} & 1.000\sd{.000} & 0.500\sd{.006} & 0.493\sd{.008} & 0.502\sd{.002} \\
 & \textsc{HOMA-add} & 1.000\sd{.000} & 1.000\sd{.000} & 1.000\sd{.000} & 0.942\sd{.065} & 0.884\sd{.064} & 0.846\sd{.087} \\
 & \textsc{HOMA} & 1.000\sd{.000} & 1.000\sd{.000} & 1.000\sd{.000} & 0.911\sd{.081} & 0.889\sd{.175} & 0.901\sd{.091} \\
\bottomrule
\end{tabular}
\end{table}

\begin{table}[!h]
\centering
\footnotesize
\setlength{\tabcolsep}{4pt}
\caption{\textbf{Accuracy by reach and depth for $w=3$ ($R_w=1$).}
Dedicated coverage-sweep runs use $d_{\mathrm{model}}=32$ and sequence length
$L=24$. Mean final test accuracy $\pm$ standard deviation over three seeds.}
\label{tab:app_coverage_depth}
\begin{tabular}{@{}ll ccc@{}}
\toprule
& & \multicolumn{3}{c}{attention layers} \\
\cmidrule(l){3-5}
reach & mechanism & $1$ & $2$ & $4$ \\
\midrule
$R{=}1$ & \textsc{Blockwise-3D} & 1.000\sd{.000} & 1.000\sd{.000} & 1.000\sd{.000} \\
 & \textsc{HOMA-add} & 1.000\sd{.000} & 1.000\sd{.000} & 1.000\sd{.000} \\
 & \textsc{HOMA} & 1.000\sd{.000} & 1.000\sd{.000} & 1.000\sd{.000} \\
\addlinespace[3pt]
$R{=}2$ & \textsc{Blockwise-3D} & 0.501\sd{.009} & 1.000\sd{.000} & 1.000\sd{.000} \\
 & \textsc{HOMA-add} & 0.858\sd{.065} & 1.000\sd{.000} & 1.000\sd{.000} \\
 & \textsc{HOMA} & 0.964\sd{.015} & 0.972\sd{.045} & 1.000\sd{.000} \\
\addlinespace[3pt]
$R{=}3$ & \textsc{Blockwise-3D} & 0.496\sd{.008} & 0.499\sd{.004} & 0.811\sd{.065} \\
 & \textsc{HOMA-add} & 0.936\sd{.071} & 0.964\sd{.040} & 0.975\sd{.044} \\
 & \textsc{HOMA} & 0.968\sd{.031} & 0.903\sd{.046} & 1.000\sd{.000} \\
\bottomrule
\end{tabular}
\end{table}

\FloatBarrier
\section{Code package and reproducibility}
\label{app:code}

The anonymized package contains the model implementations, the diagnostic-task
generators, the TAPE data pipelines, and one training script per set of runs,
with its protocol fixed in the script. Each script prints its tables from the
results it writes.

\paragraph{Seeds.}
Standard deviations use the sample ($n-1$) definition. Before aggregation, 72
duplicated \textsc{parity} records at $k=3$ and
$d_{\mathrm{model}}\in\{32,64\}$ are removed.

\end{document}